\pdfoutput=1
\documentclass{article}
\PassOptionsToPackage{svgnames,dvipsnames,table}{xcolor}
\usepackage{iclr2026_conference,times}
\usepackage{hyperref}

\usepackage{amsmath,amsfonts,bm}

\def\eqref#1{equation~\ref{#1}}

\def\1{\bm{1}}

\DeclareMathAlphabet{\mathsfit}{\encodingdefault}{\sfdefault}{m}{sl}
\SetMathAlphabet{\mathsfit}{bold}{\encodingdefault}{\sfdefault}{bx}{n}

\usepackage{url}
\usepackage{booktabs}
\usepackage{amsfonts}
\usepackage{nicefrac}
\usepackage{microtype}
\usepackage{xcolor}
\usepackage{colortbl}
\usepackage{amsmath}
\usepackage{amssymb}
\usepackage{graphicx}
\usepackage{wrapfig}
\usepackage{enumitem}
\usepackage{caption}
\usepackage[most]{tcolorbox}
\usepackage{tabularx}
\usepackage{amsthm}
\usepackage{bm}
\usepackage{multirow}
\usepackage{tikz}
\usepackage{marvosym}
\usepackage{subcaption}
\usepackage{threeparttable}
\usepackage[fixed]{fontawesome5}
\usepackage[framemethod=TikZ]{mdframed}
\usepackage{xpatch}
\usepackage{thmtools}
\usepackage{listings}

\tcbuselibrary{breakable}
\usetikzlibrary{shadows}

\definecolor{mintblue}{RGB}{210,235,250}
\definecolor{mintframe}{RGB}{120,180,220} 
\definecolor{minttitle}{RGB}{100,150,200} 
\definecolor{minttext}{RGB}{50,80,120}    

\definecolor{runzhemilk}{RGB}{255,235,245} 
\definecolor{roseframe}{RGB}{230,120,150}  
\definecolor{runzhecotton}{RGB}{255,170,200}    

\newtcolorbox{promptbox}[1]{
  enhanced,
  breakable,
  colback= runzhemilk!30!white,   
  colframe=roseframe,                
  colbacktitle= runzhecotton!66!white, 
  coltitle=white!33,
  title=\textbf{#1},
  fonttitle=\bfseries,
  sharp corners=south, 
  borderline={0.8pt}{0pt}{roseframe},
  boxrule=0.8pt,
  arc=6pt, 
  left=6pt, right=6pt, top=6pt, bottom=6pt,
  before skip=10pt, after skip=10pt,
  drop shadow=black!12,      
}

\newtcolorbox{casebox}[1]{
enhanced,
breakable,
colback=mintblue!40!white,
colframe=mintframe,
colbacktitle=minttitle!70!white,
coltitle=white,
title=\textbf{#1},
fonttitle=\bfseries,
sharp corners=south, 
borderline={0.8pt}{0pt}{minttitle},
boxrule=0.8pt,
arc=6pt, 
left=6pt, right=6pt, top=6pt, bottom=6pt,
before skip=10pt, after skip=10pt,
drop shadow=black!15, 
}

\newtcolorbox{takeawaysbox}{
enhanced,
breakable,
colback=mintblue!40!white,
colframe=mintframe,
colbacktitle=minttitle!70!white,
coltitle=white,
title=\textbf{Key Takeaways},
fonttitle=\bfseries,
sharp corners=south, 
borderline={0.8pt}{0pt}{minttitle},
boxrule=0.8pt,
arc=6pt, 
left=6pt, right=6pt, top=6pt, bottom=6pt,
before skip=10pt, after skip=10pt,
drop shadow=black!15, 
}

\makeatletter
\newcommand{\DrawLine}{%
  \begin{tikzpicture}
  \path[use as bounding box] (0,0) -- (\linewidth,0);
  \draw[color=minttitle!70!white,dashed,dash phase=1.5pt]
        (0-\kvtcb@leftlower-\kvtcb@boxsep,0)--
        (\linewidth+\kvtcb@rightlower+\kvtcb@boxsep,0);
  \end{tikzpicture}%
  }
\makeatother

\mdfdefinestyle{mdpurpleremark}{%
  skipabove=8pt,
  skipbelow=8pt,
  linewidth=3pt,
  rightline=false,
  leftline=true,
  topline=false,
  bottomline=false,
  linecolor=Violet!50!black,
  backgroundcolor=Orchid!5,
  innertopmargin=6pt,
  innerbottommargin=6pt,
  innerleftmargin=8pt,
  innerrightmargin=8pt,
}

\declaretheoremstyle[
  headfont=\bfseries\color{Violet!50!black},
  bodyfont=\normalfont,
  spaceabove=0pt,
  spacebelow=0pt,
  mdframed={style=mdpurpleremark}
]{purpleremark}

\declaretheorem[style=purpleremark,name=Remark,numbered=no]{remark*}

\makeatletter
\xpatchcmd{\endmdframed}
  {\aftergroup\endmdf@trivlist\color@endgroup}
  {\endmdf@trivlist\color@endgroup\@doendpe}
  {}{}
\makeatother

\mdfdefinestyle{mdpurplebox}{%
  roundcorner=10pt,
  linewidth=1pt,
  skipabove=12pt,
  skipbelow=12pt,
  innertopmargin=9pt,
  innerbottommargin=9pt,
  linecolor=black,
  nobreak=true,
  backgroundcolor=Orchid!10,
  shadow=true,
  shadowsize=6pt,
  shadowcolor=black!30,
  frametitleaboveskip=8pt,
  frametitlebelowskip=8pt,
  frametitlebackgroundcolor=Violet!50!black,
  frametitlefont=\bfseries\sffamily\color{white},
  frametitlerule=true,
}

\definecolor{darkblue}{rgb}{0, 0, 0.5}
\hypersetup{colorlinks=true, citecolor=darkblue, urlcolor=darkblue, linkcolor=darkblue}

\lstdefinestyle{combcode}{
  basicstyle=\tiny\ttfamily,
  columns=fullflexible,
  breaklines=true,
  breakatwhitespace=false,
  frame=single,
  rulecolor=\color{orange!55!black},
  backgroundcolor=\color{orange!8},
  xleftmargin=2pt,
  xrightmargin=2pt,
  aboveskip=4pt,
  belowskip=4pt,
  captionpos=b
}

\lstdefinestyle{combpayload}{
  basicstyle=\footnotesize\ttfamily,
  columns=fullflexible,
  breaklines=true,
  breakatwhitespace=false,
  frame=single,
  rulecolor=\color{orange!55!black},
  backgroundcolor=\color{orange!8},
  xleftmargin=2pt,
  xrightmargin=2pt,
  aboveskip=4pt,
  belowskip=4pt,
  captionpos=b
}

\newcommand{\projectpage}{\raisebox{-1.5pt}{\faIcon{globe}}}
\newcommand{\github}{\raisebox{-1.5pt}{\includegraphics[height=1.05em]{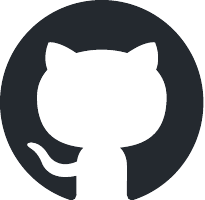}}}

\NewDocumentCommand{\haoran}{mO{}}{\textcolor{blue}{\textsuperscript{\textit{haoran}}\textsf{\textbf{\small[#1]}}}}
\NewDocumentCommand{\runzhe}{mO{}}{\textcolor{blue}{\textsuperscript{\textit{runzhe}}\textsf{\textbf{\small[#1]}}}}
\NewDocumentCommand{\yizhuo}{mO{}}{\textcolor{blue}{\textsuperscript{\textit{yizhuo}}\textsf{\textbf{\small[#1]}}}}
\NewDocumentCommand{\yafu}{mO{}}{\textcolor{blue}{\textsuperscript{\textit{yafu}}\textsf{\textbf{\small[#1]}}}}

\usepackage[capitalize,noabbrev]{cleveref}

\makeatletter
\providecommand{\@LN@col}[1]{}
\providecommand{\@LN}[2]{}
\makeatother

\title{ComBench: A Benchmark for Rigorous Proof Reasoning and Constructive Realization in Olympiad-Level Combinatorics}

\newcommand{\authormark}[1]{\textsuperscript{#1}}
\newcommand{\equalmark}{\textsuperscript{*}}
\newcommand{\corrsym}{\ensuremath{\dagger}}
\newcommand{\corrmark}{\textsuperscript{\corrsym}}

\author{%
\small
{\bfseries Shunkai Zhang\authormark{2,1,*} \quad
Haoran Zhang\authormark{3,1,*} \quad
Yun Luo\authormark{1,\corrsym} \quad
Qianjia Cheng\authormark{1}} \\
{\bfseries Haodi Lei\authormark{1} \quad
Yizhuo Li\authormark{1} \quad
Runzhe Zhan\authormark{1} \quad
Zhilin Wang\authormark{1} \quad
Bangjie Xu\authormark{4}} \\
{\bfseries Yucheng Su\authormark{4} \quad
Xinmiao Han\authormark{4} \quad
Xiaoye Qu\authormark{1} \quad
Dongrui Liu\authormark{1} \quad
Zhouchen Lin\authormark{2}} \\
{\bfseries Yu Qiao\authormark{1} \quad
Ning Ding\authormark{4, 1} \quad
Yafu Li\authormark{1, 5,\corrsym} \quad
Yu Cheng\authormark{1,\corrsym}} \\
\normalfont\small \authormark{1}Shanghai AI Laboratory \quad
\authormark{2}Peking University \quad
\authormark{3}Shanghai Jiao Tong University \\
\normalfont\small \authormark{4}Tsinghua University \quad
\authormark{5}The Chinese University of Hong Kong \\
\normalfont\small
\equalmark Equal contribution. \quad
\corrmark Corresponding authors.
}

\iclrfinalcopy

\begin{document}

\maketitle
\vspace{-1.45em}
\begin{abstract}

Combinatorics is central to Olympiad-level mathematical problem solving, requiring deep discrete reasoning, creative constructions, and rigorous structural insight. Recent evidence suggests that even today's strongest frontier models remain uneven on Olympiad combinatorics, revealing a gap in creative mathematical reasoning. We introduce ComBench, an Olympiad-level combinatorics benchmark for evaluating and diagnosing the combinatorial reasoning capabilities of large language models. ComBench contains 100 human-annotated competition-level problems organized around two complementary settings: \textit{analysis-centric problems}, which primarily require rigorous mathematical arguments, and \textit{construction-centric problems}, which require explicit constructions in addition to correctness justifications. The evaluation protocol combines rubric-guided proof grading with deterministic construction verification, exposing cases where proof quality and construction validity diverge. Experiments on frontier open- and closed-source models show that ComBench is far from saturated: the strongest model reaches \(65.4\%\) overall Avg. and \(75.3\%\) overall Best@4. We further find that \textit{Rigorous Proof Reasoning} and \textit{Constructive Realization} are distinct capabilities: Kimi-K2.6 trails GPT-5.5 on analysis-centric proof grading but surpasses it on construction-centric Best@4, while \textit{Existence and Construction} problems remain consistently hardest across representative frontier models.

\end{abstract}

\begin{center}
\vspace{-1em}
~\projectpage~\href{https://simplified-reasoning.github.io/ComBench/docs/}{{\text{Project Page}}}
\quad \quad \quad
~\github~\href{https://github.com/Simplified-Reasoning/ComBench}{{\text{Code}}}
\end{center}

\bibliographystyle{iclr2026_conference}
\section{Introduction}

Large language models (LLMs) have recently demonstrated rapid progress on mathematical reasoning, with frontier systems approaching elite human performance on several competition-style benchmarks \citep{2025robustmathematicalreasoning,2025amobench}. Nevertheless, their success remains uneven on Olympiad-level combinatorics, a domain that requires not only long-horizon logical deduction but also discrete structural insight and creative construction. For example, Gemini Deep Think and DeepSeekMath‑V2\citep{2025deepseekmathv2} achieved gold-level performance at IMO 2025, solving five of six problems, yet failed on the hardest combinatorics problem \citep{2025robustmathematicalreasoning}. It suggests that current models may still lack key ingredients of constructive and creative mathematical reasoning. 

\begin{figure}[t]
  \centering
  \includegraphics[width=\linewidth]{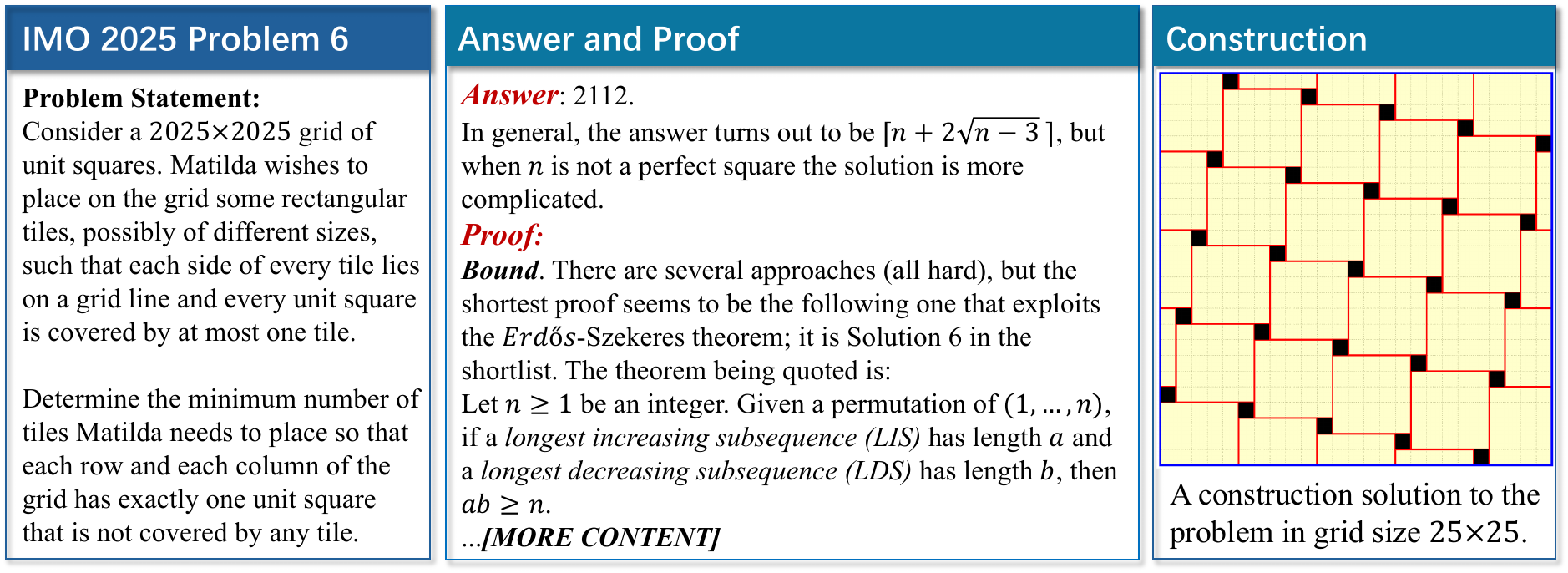}
  \caption{IMO 2025 P6, a challenging combinatorics problem unsolved by all evaluated models. The figure shows the original problem statement, the reference answer, and a schematic illustration of the reference construction.}
  \label{fig:imo26_6}
\end{figure}


Existing Olympiad-level benchmarks \citep{2025robustmathematicalreasoning,2025amobench} provide important tools for studying such progress, but they do not isolate the constructive nature of combinatorics. Answer-based benchmarks can scale evaluation, but reveal little about whether a model has produced a valid mathematical argument. IMO-Bench-style proof evaluation addresses this limitation by judging proof quality and partial progress, making it well-suited for measuring \textit{Rigorous Proof Reasoning}: the ability to write complete solutions aligned with Olympiad grading standards. 
However, many combinatorial problems require an additional capability. A model must not only argue that an object exists, but also realize it explicitly.

We refer to this second capability as \textit{Constructive Realization}: the ability to turn an existential or constructive argument into a complete, unambiguous, and machine-checkable discrete witness. Such witnesses may be colorings, tilings, graphs, set families, strategies, matrices, permutations, or counterexamples. Natural-language descriptions of these objects are often difficult to evaluate reliably, since their validity depends on global constraints over the entire object. For example, in IMO 2025 Problem 6 (Figure~\ref{fig:imo26_6}), a complete solution requires not only proving the optimal value, but also exhibiting a tile distribution that attains it. 

In this study, we propose \textbf{ComBench}, a benchmark of 100 Olympiad-level combinatorics problems designed to evaluate both \textit{Rigorous Proof Reasoning} and \textit{Constructive Realization}. ComBench extends IMO-Bench-style proof evaluation with executable construction verification. It contains two types of records. \textit{Analysis-centric} records evaluate the original mathematical solution using rubric-guided proof grading. \textit{Construction-centric} records retain the original proof task but add a construction task, where the model must output a concrete witness in a prescribed representation, and an item-specific verifier checks the witness deterministically. The distinction lies in the evaluation target.

ComBench is curated from major Olympiad-style competitions, including IMO, USAMO, team selection tests, regional Olympiads, and IMO Shortlist problems. Each problem is annotated with grading guidelines following a \(0/1/6/7\)-point rubric. Construction-centric records are further equipped with construction instructions, reference witnesses, verifier plans, and deterministic Python verifiers. 
Multiple math experts provide rigorous and carefully justified annotations concerning grading guidelines, combinatorial construction instructions, and deterministic verifiers.
This design lets us evaluate proof quality and construction validity separately, and then combine them through verifier-gated scoring when a construction is required.


We evaluate a range of frontier closed-source and open-source LLMs on ComBench, with several findings illustrating the benchmark's diagnostic value. Overall, Olympiad-level combinatorics remains far from saturated: the strongest model, GPT-5.5, achieves \(65.4\%\) overall Avg. and \(75.3\%\) overall Best@4 under our rubric-guided and verifier-gated evaluation. Model behavior also differs sharply across the two target capabilities. Kimi-K2.6 is substantially weaker than GPT-5.5 on analysis-centric proof grading, yet achieves the best construction-centric Best@4 score, surpassing GPT-5.5. At the category level, \textit{Existence and Construction} problems are consistently the hardest for representative frontier models, while \textit{Counting} and \textit{Graph Theory} problems are comparatively easier. These findings suggest that \textit{Rigorous Proof Reasoning} and \textit{Constructive Realization} are related but distinct capabilities, and that progress on Olympiad-level combinatorics requires both proof-level rigor and explicit witness realization.

\paragraph{Contributions.}
This work makes the following contributions: 
    \begin{itemize}[left=0pt]
         \item We introduce \textbf{ComBench}, a 100-problem Olympiad-level combinatorics benchmark for \textit{Rigorous Proof Reasoning} and \textit{Constructive Realization}.

        \item We develop an evaluation protocol that combines IMO-Bench-style rubric-guided proof grading with deterministic verification of explicit construction witnesses.
        
        \item We provide a systematic empirical evaluation of frontier closed-source and open-source LLMs on ComBench, revealing a clear separation between proof-level reasoning and explicit construction ability, as well as category-specific weaknesses in Olympiad-level combinatorics.
    \end{itemize}

\section{Related Work}

\paragraph{Evaluation of Mathematical Reasoning.}
Mathematical benchmarks have evolved from GSM8K and MATH \citep{2021trainingverifiers,2021measuringmathematicalproblemsolving} to Olympiad- and frontier-level evaluations \citep{2024olympiadbench,2026matharena,2025frontiermathbenchmark,2025hle,2026horizonmath}, dominated by three protocols: answer-only matching as in AMO-Bench and OlymMATH \citep{2025amobench,2025olymmath}, rubric-guided $0$--$7$ proof grading as in IMO-Bench and ProofGrader \citep{2025robustmathematicalreasoning,2025proofbench}, and Lean-based formal proving benchmarks such as CombiBench \citep{2025combibench}, which formalizes combinatorial problems in Lean 4 and evaluates theorem-proving and fill-in-the-blank formal solutions. Yet Olympiad combinatorics, where solutions hinge on globally constrained witnesses such as colorings, set families, or counter-examples, remains underserved by these protocols: answer matching cannot expose invalid witnesses, and proof graders rate the prose without checking the construction itself. Complementing the Lean-formalization line, ComBench retains natural-language solutions but adds executable verification of explicit witnesses.

\paragraph{LLMs for Olympiad-level Reasoning.}
After early neuro-symbolic and formal-proof systems reached medal-level IMO performance \citep{2025alphageometry2,2025alphaproof}, large-scale post-training with reinforcement learning, further refined by recipes around scaling, training strategies, and tool-integrated reasoning, has substantially lifted the mathematical capabilities of general-purpose LLMs \citep{2025deepseekr1,2025kimik2,2025qwen3,cheng2026teachingthinkingmodelsreason,zhang2026characterizing}. Building on this trajectory, recent thinking systems have reached IMO-gold-level natural-language performance under contest conditions \citep{2025robustmathematicalreasoning,2025geminiimohuangyang,chen2025seedprover15masteringundergraduatelevel,Nemotron_Cascade_2,SU2026}, yet the hardest combinatorics problems remain unsolved: no model received credit on IMO 2025 P6. ComBench is designed to disentangle and measure precisely this gap.

\section{ComBench Dataset}
\label{sec:dataset}

\subsection{Overview}


\begin{wrapfigure}{r}{0.38\textwidth}
    \vspace{-1.0em}
    \centering
    \captionsetup{font=small,skip=2pt}
    \includegraphics[width=0.92\linewidth]{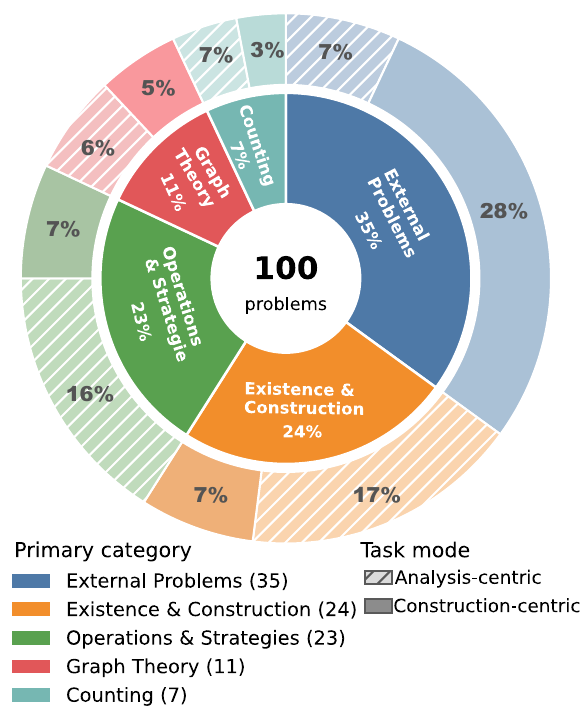}
    \caption{Distribution of ComBench categories.}
    \label{fig:category-distribution}
    \vspace{-1.1em}
\end{wrapfigure}


We introduce ComBench, a benchmark of 100 Olympiad-level combinatorics problems designed to evaluate mathematical reasoning beyond final-answer correctness. It is balanced between 50 analysis-centric problems and 50 construction-centric problems. The benchmark spans 15 competition sources from 2000 to 2025, including the International Mathematical Olympiad (IMO), United States of America Mathematical Olympiad (USAMO), team selection tests, regional Olympiads, and IMO Shortlist problems. This source diversity combines canonical international problems with national and selection-test problems that broaden coverage of constructive, extremal, graph-theoretic, counting, and strategy-driven combinatorics.
Figure~\ref{fig:category-distribution} visualizes this composition: the 100 problems are split evenly between analysis-centric and construction-centric records, and are organized into five primary combinatorics categories.

The two problem types operationalize the paper's two target capabilities: analysis-centric records evaluate \textit{Rigorous Proof Reasoning}, while construction-centric records additionally evaluate \textit{Constructive Realization} through an explicit, mechanically checkable witness. Aggregate source, year, category, and problem-type statistics are provided in Appendix~\ref{app:dataset-statistics}.


\begin{figure}[t]
\centering
\includegraphics[width=\linewidth]{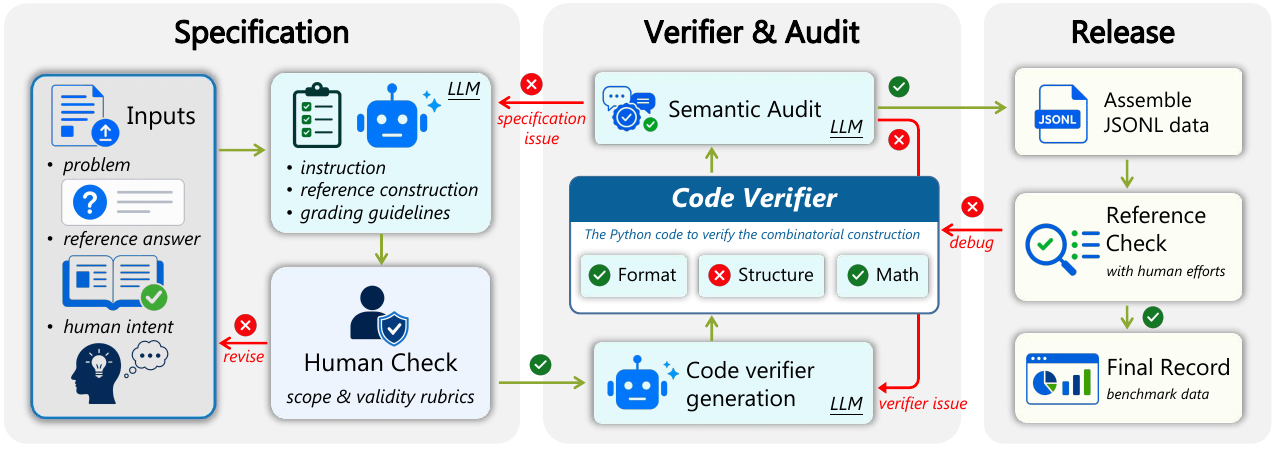}
\caption{Construction-centric data builder pipeline for ComBench.}
\label{fig:combench-construction-pipeline}
\end{figure}

\subsection{Record Design and Problem Taxonomy}

We select Olympiad-level combinatorics problems centered on discrete structures, extremal reasoning, counting, graph-theoretic arguments, strategies, or explicit constructions.
The selected problems include standard Olympiad-style proof tasks as well as tasks whose natural solutions require explicit witnesses such as colorings, permutations, graphs, tables, set systems, tilings, strategies, or counterexamples. To better target combinatorial reasoning, we exclude problems whose core difficulty is non-combinatorial, even if they involve minor counting or discrete elements.


Each ComBench record contains the information needed for rubric-guided proof evaluation: the original problem statement, source metadata, reference answer, reference solution when available, primary problem category, problem type, and problem-specific grading guidelines. The grading guidelines follow a \(0/1/6/7\)-point rubric and specify what counts as no progress, minimal progress, nearly complete progress, and a complete solution. Construction-centric records contain additional fields that define the executable construction task: an item-specific construction instruction, a reference witness, an informal verifier plan, and deterministic Python verifier code. Appendix~\ref{app:jsonl-example} gives a complete formatted example of one construction-centric record.

For each construction-centric problem, the \textit{witness} is the concrete discrete object requested by the construction instruction. It is not a proof sketch or natural-language justification; rather, it is the object whose representation and mathematical validity can be checked by a verifier. This design makes the construction task explicit, rather than relying on free-form object descriptions that may be difficult to parse or reproduce. Representative witness formats are shown in Appendix~\ref{app:witness-forms}.

ComBench uses five primary problem categories: extremal problems, existence and construction, operations and strategies, graph theory, and counting. These categories provide a stable way to analyze recurring reasoning patterns rather than mutually exclusive mathematical subfields. Category definitions and the category-by-type breakdown are provided in Appendix~\ref{app:dataset-statistics}.

\subsection{Annotation Pipeline}

Both analysis-centric and construction-centric records require annotation. For analysis-centric records, the main annotation is the problem-specific \(0/1/6/7\)-point grading guideline used for rubric-guided proof evaluation. Construction-centric records use the same proof-rubric annotation, but additionally require a construction instruction, a reference witness, an informal verifier plan, deterministic verifier code, and semantic audit. Because rubric construction is shared by both record types, we focus below on the full construction-centric pipeline, which extends proof-rubric annotation with witness specification and executable verification, as shown in Figure~\ref{fig:combench-construction-pipeline}.


\paragraph{Stage 1: specification and rubric construction.}
The inputs include the original problem, reference answer, and human-expert intent for the customized construction task, together with a reference witness. The intent specifies what kind of witness should be requested and what mathematical target it must realize. Given these inputs, the LLM drafts an item-specific construction instruction, standardizes the reference construction into the required format, and produces problem-specific grading guidelines under the \(0/1/6/7\)-point rubric. Human reviewers then check whether the construction instruction preserves the original problem meaning, whether the reference construction satisfies the intended target, and whether the grading guideline captures meaningful Olympiad-style proof progress.

\paragraph{Stage 2: verifier generation and semantic audit.}
After human review, the problem statement and construction instruction are used to generate deterministic Python verifier code. The verifier reads only the raw construction payload and checks both the declared representation and the required mathematical constraints. We then use an LLM-assisted semantic audit to inspect the full chain from the original problem to the executable verifier. The audit checks whether the construction target, instruction, witness, verifier plan, and code remain aligned with the original mathematical goal, and flags either an upstream specification issue or a verifier-generation issue when they diverge.

\paragraph{Stage 3: record assembly and executable reference check.}
Records passing the semantic audit are assembled into JSONL format. As an executable reference check, the reference witness is passed to the verifier through the same raw-standard-input interface used for model submissions. A record is accepted only if this check succeeds. It then receives a final human inspection of verifier strictness, coverage, consistency, and diagnostic clarity. The full record schema, verifier I/O contract, and human audit checklist are described in Appendix~\ref{app:annotation-pipeline},~\ref{app:quality-control}, and~\ref{app:verifier-details}.

\subsection{Quality Control and Transparency}

Quality control in ComBench targets three main risks. First, we check \textit{target fidelity}: the problem statement, reference answer, grading guideline, construction instruction, and reference witness must refer to the same mathematical claim. Second, we check \textit{verifier adequacy}: each reference witness must be parseable and accepted by its verifier, and each verifier must enforce the intended global combinatorial constraints rather than only superficial format properties. Third, we check \textit{metadata and evaluation consistency}: source identifiers, categories, problem types, record counts, and final JSONL files must match the files used by the evaluation pipeline. Additional record-schema and quality-control details are provided in Appendix~\ref{app:quality-control}.

We also check the overlap between ComBench and IMO-Bench. We compare all 100 problems using source metadata, problem-statement similarity, and manual inspection of high-similarity candidates. This process identifies 14 overlapping or substantially modified problems, yielding an overlap rate of 14\%; the remaining 86 problems show no high-confidence match. 
The matching protocol and released overlap metadata are described in Appendix~\ref{app:imobench-overlap}.

\begin{wrapfigure}{r}{0.34\textwidth}
 \vspace{-5.5em}
 \centering
 \captionsetup{font=small,skip=2pt}
 \includegraphics[width=\linewidth]{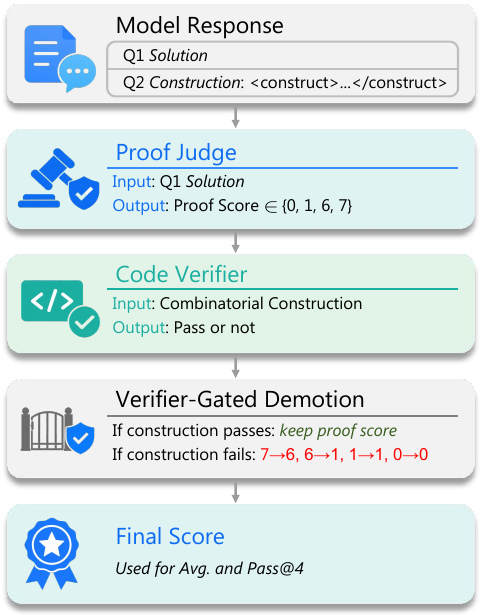}
 \caption{Verifier-gated scoring rule for construction-centric records.}
 \label{fig:verifier-gated-scoring}
 \vspace{-1.0em}
\end{wrapfigure}

\subsection{Combining Proof and Construction Scores}
\label{sec:score-integration}

ComBench keeps proof grading and construction verification as separate signals so that \textit{Rigorous Proof Reasoning} and \textit{Constructive Realization} can be analyzed independently. For model ranking, however, construction-centric records also require a single score. Analysis-centric records use the proof score returned by the rubric-guided judge. For construction-centric records, we obtain a proof score \(p \in \{0,1,6,7\}\) and a binary construction score \(c \in \{0,1\}\). If construction passes, the final score remains \(p\). If construction fails, high proof scores are demoted: \(7\mapsto 6\) and \(6\mapsto 1\), while \(0\) and \(1\) remain unchanged.

Figure~\ref{fig:verifier-gated-scoring} illustrates this verifier-gated rule. The rule is motivated by manual inspection of construction-centric responses: when a high-scoring proof fails executable construction verification, the proof often contains a hidden construction gap, such as asserting existence at a high level, leaving the witness under-specified, or failing to instantiate the object in a checkable form. Thus a valid witness is not treated as an extra bonus; it confirms that a high-scoring constructive solution can actually be realized. Conversely, failed construction should not erase all proof progress, since the response may still contain meaningful mathematical reasoning. The demotion rule therefore calibrates high proof scores using construction validity while preserving low proof scores that already indicate insufficient progress. Appendix~\ref{app:case-study} provides examples of this scoring rationale.

 \section{Experiments}
\label{sec:experiments}

The experiments evaluate what current models can and cannot do under the two capabilities introduced earlier: \textit{Rigorous Proof Reasoning} and \textit{Constructive Realization}. We first report overall performance on ComBench, then examine whether success on proof reasoning transfers to construction realization, which combinatorial categories remain difficult, and what mathematical proof errors dominate below-full-credit solutions.

\subsection{Evaluation Configuration}

\paragraph{Models and sampling.}
We evaluate ten open- and closed-source models: GPT-5.5~\citep{openai_gpt55_2026}, Gemini-3.1-Pro-Preview-Thinking~\citep{gemini31pro_2026}, Kimi-K2.6~\citep{kimi_k26_2026}, DeepSeek-V4-Pro~\citep{deepseekai2026deepseekv4}, Nemotron-Cascade-2-\allowbreak 30B-A3B~\citep{2026nemotron}, GLM-5.1~\citep{glm5team2026glm5vibecodingagentic}, Qwen3.6-35B-A3B~\citep{qwen36_35b_a3b}, Qwen3.6-Max-Preview~\citep{qwen36_max_preview}, Google-Gemma-4-31B-IT~\citep{farabet2026gemma4}, and SU-01\citep{SU2026}. We use temperature \(0.6\), model-specific maximum token limits, and four sampled solutions per item. 

\vspace{-2mm}
\paragraph{Prompt format.}
The prompt follows the record type. Analysis-centric records ask for a complete Olympiad-style solution to the original problem. Construction-centric records contain two questions: the original proof task and an item-specific construction task. The construction answer must provide one \texttt{<construct>...</construct>} payload, read by the executable verifier. Representative prompt templates are provided in Appendix~\ref{app:prompt-templates}.

\vspace{-2mm}
\paragraph{Evaluation components.}
Proofs are scored by a rubric-guided LLM judge using the problem statement, reference solution when available, and problem-specific grading guidelines. The judge returns one of \(0,1,6,7\) points out of \(7\). For construction-centric records, the extracted construction payload is additionally passed to the item-specific Python verifier and receives a binary pass/fail result. The verifier-gated score used for construction-centric records follows Section~\ref{sec:score-integration}.
We further conduct a manual audit of sampled proof-judge decisions in Appendix~\ref{app:human-auto-alignment}.

\vspace{-2mm}
\paragraph{Metrics.}
We report \textit{Avg.}, the average normalized score over all generated solutions, and \textit{Best@4}, which takes the best score among four sampled solutions for each record before averaging across records. Avg. measures overall model capability across sampled solutions, while Best@4 measures potential performance under repeated sampling. Full prompt templates, parsing rules, evaluator details, and failure handling are provided in Appendix~\ref{app:evaluation-details}.

\subsection{Main Results}

\begin{table}[t]
\centering
\small
\resizebox{0.9\linewidth}{!}{
\begin{tabular}{lcc|cc|cc}
\toprule
\multirow{2.5}{*}{\textbf{Model}}  
& \multicolumn{2}{c|}{\textbf{Analysis-centric}} 
& \multicolumn{2}{c|}{\textbf{Construction-centric}} 
& \multicolumn{2}{c}{\textbf{Overall}} \\
\cmidrule(lr){2-3}\cmidrule(lr){4-5}\cmidrule(lr){6-7}
& \textbf{Avg.(\%)} 
& \textbf{Best@4(\%)} 
& \textbf{Avg.(\%)} 
& \textbf{Best@4(\%)} 
& \textbf{Avg.(\%)} 
& \textbf{Best@4(\%)} \\
\midrule
GPT-5.5 
& \cellcolor{blue!3}\textbf{62.4} & \cellcolor{blue!3}\textbf{72.9} 
& \cellcolor{green!3}\textbf{68.4} & \cellcolor{green!3}77.7 
& \cellcolor{orange!4}\textbf{65.4} & \cellcolor{orange!4}\textbf{75.3} \\
Gemini-3.1-Pro 
& \cellcolor{blue!3}56.1 & \cellcolor{blue!3}69.7 
& \cellcolor{green!3}64.5 & \cellcolor{green!3}78.3 
& \cellcolor{orange!4}60.3 & \cellcolor{orange!4}74.0 \\
Kimi-K2.6 
& \cellcolor{blue!3}43.5 & \cellcolor{blue!3}60.6 
& \cellcolor{green!3}63.4 & \cellcolor{green!3}\textbf{83.7} 
& \cellcolor{orange!4}53.5 & \cellcolor{orange!4}72.1 \\
DeepSeek-V4-Pro 
& \cellcolor{blue!3}37.8 & \cellcolor{blue!3}56.6 
& \cellcolor{green!3}52.6 & \cellcolor{green!3}67.7 
& \cellcolor{orange!4}45.2 & \cellcolor{orange!4}62.1 \\
Qwen3.6-Max 
& \cellcolor{blue!3}21.4 & \cellcolor{blue!3}32.9 
& \cellcolor{green!3}28.4 & \cellcolor{green!3}39.1 
& \cellcolor{orange!4}24.9 & \cellcolor{orange!4}36.0 \\
SU-01 
& \cellcolor{blue!3}20.9 & \cellcolor{blue!3}30.3 
& \cellcolor{green!3}28.8 & \cellcolor{green!3}41.1 
& \cellcolor{orange!4}24.8 & \cellcolor{orange!4}35.7 \\
GLM-5.1 
& \cellcolor{blue!3}21.6 & \cellcolor{blue!3}36.0 
& \cellcolor{green!3}25.6 & \cellcolor{green!3}37.1 
& \cellcolor{orange!4}23.6 & \cellcolor{orange!4}36.6 \\
Qwen3.6-35B 
& \cellcolor{blue!3}17.9 & \cellcolor{blue!3}26.6 
& \cellcolor{green!3}22.7 & \cellcolor{green!3}32.0 
& \cellcolor{orange!4}20.3 & \cellcolor{orange!4}29.3 \\
Nemotron-Cascade
& \cellcolor{blue!3}21.8 & \cellcolor{blue!3}32.9 
& \cellcolor{green!3}17.4 & \cellcolor{green!3}28.0 
& \cellcolor{orange!4}19.6 & \cellcolor{orange!4}30.4 \\
Gemma-4-31B-IT 
& \cellcolor{blue!3}16.1 & \cellcolor{blue!3}24.3 
& \cellcolor{green!3}17.5 & \cellcolor{green!3}30.9 
& \cellcolor{orange!4}16.8 & \cellcolor{orange!4}27.6 \\
\bottomrule
\end{tabular}
}
\caption{Main evaluation results on ComBench.}
\vspace{-1mm}
\label{tab:main-results}
\end{table}

The main results report analysis-centric, construction-centric, and overall performance; all values are percentages, and construction-centric scores use the verifier-gated rule in Section~\ref{sec:score-integration}. Table~\ref{tab:main-results} shows that ComBench is far from saturated: the strongest model, GPT-5.5, reaches only \(65.4\%\) overall Avg. and \(75.3\%\) overall Best@4, leaving substantial headroom even under best-of-four sampling. Thus Olympiad-level combinatorics remains difficult under rubric-guided and verifier-gated evaluation.

The rankings reveal both model tiers and capability separation. GPT-5.5 leads overall and on analysis-centric records, while Gemini-3.1-Pro remains close in overall Best@4. Kimi-K2.6, however, achieves the highest construction-centric Best@4 despite a lower analysis-centric score, showing that \textit{Constructive Realization} is not merely a by-product of stronger \textit{Rigorous Proof Reasoning}. DeepSeek-V4-Pro forms a middle tier below the top three but above the remaining models. The lower-scoring group, including Qwen3.6-Max, SU-01, GLM-5.1, Qwen3.6-35B, Nemotron-Cascade, and Gemma-4-31B-IT, further shows that ComBench remains challenging for smaller or more specialized models. SU-01 is especially informative: its overall score is limited by format and instruction-following failures, yet parseable responses still show nontrivial mathematical performance on construction-centric records.

These patterns motivate verifier-gated evaluation: a fluent proof can receive high natural-language credit without producing the claimed object in a checkable form. The results suggest that future mathematical models must improve not only proof fluency, but also target identification, structural planning, and explicit witness synthesis. Additional per-category and per-problem results are reported in Appendix~\ref{app:additional-results}.

\subsection{Subset Analysis}

We next separate the main results into two views: construction-centric behavior and category-level behavior. The first view tests whether proof success is accompanied by a valid witness, while the second view asks which forms of combinatorial reasoning remain most difficult.

\paragraph{Construction-centric behavior.}
\begin{wraptable}{r}{0.48\textwidth}
\vspace{-1.0em}
\centering
\scriptsize
\captionsetup{font=small,skip=2pt}
\resizebox{\linewidth}{!}{
\begin{tabular}{lccc}
\toprule
\textbf{Model} & \textbf{Proof} & \textbf{Construction} & \textbf{Score} \\
\midrule
GPT-5.5 & \textbf{72.3} & \textbf{75.0} & \textbf{68.4} \\
Gemini-3.1-Pro & 68.9 & \textbf{75.0} & 64.5 \\
Kimi-K2.6 & 66.3 & \textbf{75.0} & 63.4 \\
DeepSeek-V4-Pro & 56.1 & 62.5 & 52.6 \\
SU-01 & 29.6 & 27.5 & 28.8 \\
Qwen3.6-Max & 31.5 & 38.0 & 28.4 \\
GLM-5.1 & 29.0 & 27.0 & 25.6 \\
Qwen3.6-35B & 24.9 & 40.5 & 22.7 \\
Gemma-4-31B-IT & 20.9 & 15.0 & 17.5 \\
Nemotron-Cascade & 18.4 & 20.5 & 17.4 \\
\bottomrule
\end{tabular}
}
\caption{Construction-centric subset analysis. All values are average percentages.}
\label{tab:construction-subset-analysis}
\vspace{0.4em}
\captionsetup{type=figure,font=small,skip=2pt}
\includegraphics[width=\linewidth]{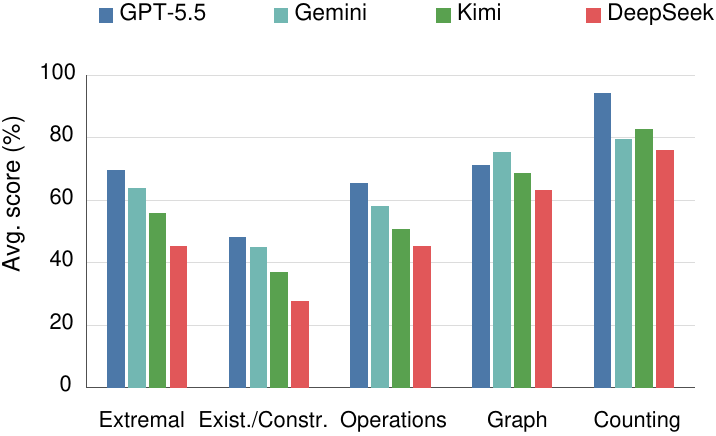}
\captionof{figure}{Category-level performance of four representative frontier models.}
\label{fig:category-performance-top4}
\vspace{-1.0em}
\end{wraptable}

Table~\ref{tab:construction-subset-analysis} focuses on the 50 construction-centric records, with all values reported as percentages. The proof-score column measures \textit{Rigorous Proof Reasoning}, the construction pass rate measures \textit{Constructive Realization}, and the verifier-gated score measures whether the two are jointly present. Across models, verifier-gated scores are lower than proof scores, showing that high-scoring proofs are not always accompanied by valid executable constructions.

The construction pass rate further isolates witness realization in the prescribed representation. The strongest models reach a \(75.0\%\) per-sample construction pass rate, still leaving many failed witnesses even among frontier systems. Kimi-K2.6 matches the top construction pass rate and achieves a strong verifier-gated score, despite weaker analysis-centric performance in Table~\ref{tab:main-results}. This reinforces the main-results observation that construction-centric records test more than proof difficulty: they require translating a mathematical target into a complete discrete object satisfying global constraints. Appendix~\ref{app:case-study} gives two GPT-5.5 examples where high proof scores are demoted because the response lacks an accepted mechanically checkable witness.

\begin{figure}[t]
\centering
\includegraphics[width=0.96\textwidth]{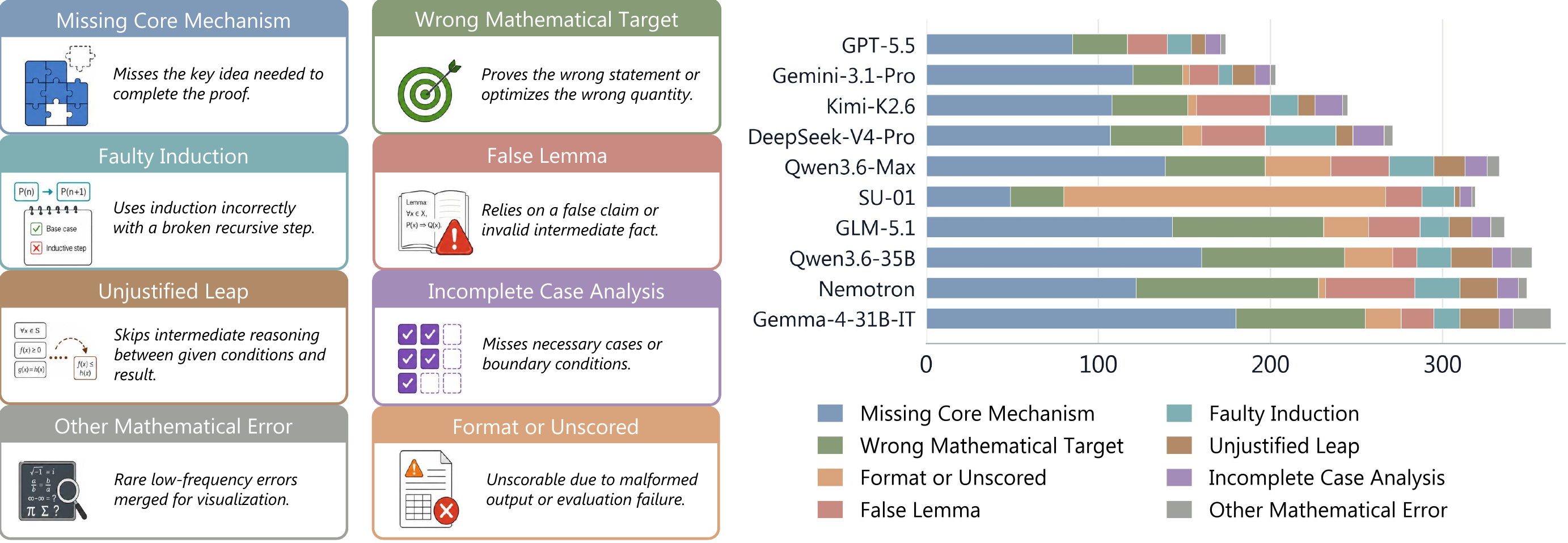}
\caption{Distribution of primary proof error reasons over all below-full-credit proof samples. Primary reasons are mutually exclusive by construction, while secondary reason tags may overlap.}
\label{fig:error-reason-distribution}
\end{figure}

\paragraph{Category-level behavior.}

Figure~\ref{fig:category-performance-top4} breaks down category-level performance for the four strongest representative models. Analysis-centric records use proof scores, while construction-centric records use verifier-gated scores. Across all four models, \textit{Existence and Construction} is the hardest category, with scores below the corresponding model's performance on the other categories. 
This suggests that current LLMs remain weak at existence-style combinatorial reasoning: they often struggle to identify the right global object, maintain all constraints simultaneously, and turn an existence argument into a complete witness.

Counting and Graph Theory obtain the highest scores, indicating that current models are relatively stronger when the solution can be organized around enumerative structure, graph reformulation, or local structural constraints. By contrast, Existence and Construction, Operations and Strategies, and Extremal Problems often require selecting the correct global target, maintaining an invariant across a process, or matching a sharp bound with a concrete construction.
This imbalance suggests that models have acquired some local structural reasoning, but remain weak at global combinatorial planning: they can often manipulate a given representation, yet struggle to choose the right object, invariant, or extremal structure before the proof begins.

\subsection{Error Analysis}

This section analyzes proof-side failures only. A generated solution is included if its proof score is \(0\), \(1\), or \(6\), or if the proof component is unscored because of parsing or evaluation failure. Construction failure is analyzed separately through the construction pass rate and verifier-gated score in Table~\ref{tab:construction-subset-analysis}. The two analyses are complementary: proof errors diagnose failures in \textit{Rigorous Proof Reasoning}, while construction failures diagnose whether the model can realize the required witness in a complete and checkable form.

We assign one primary reason to each below-full-credit proof critique using mathematically grounded proof-error types. Each case is first annotated with a fine-grained category, and mathematical categories below the global \(2\%\) frequency threshold are merged into \textit{Other Mathematical Error} for the paper-facing distribution. Primary reasons are mutually exclusive by construction, while secondary tags may overlap. Figure~\ref{fig:error-reason-distribution} shows that most failures are not merely formatting errors. The dominant failure mode is \textit{Missing Core Mechanism}, accounting for \(41.2\%\) of below-full-credit proof solutions. A further \(20.0\%\) fall under \textit{Wrong Mathematical Target}, where the model proves or optimizes the wrong statement. \textit{Format or Unscored} accounts for \(11.0\%\), while false lemmas, faulty induction, unjustified leaps, and incomplete case analysis together account for \(25.5\%\). Full definitions, fine-grained counts, and per-model error distributions are provided in Appendix~\ref{app:error-taxonomy}.

These failures indicate that current models are not limited only by response formatting or proof presentation.
The dominant errors point to deeper weaknesses in global proof planning: models often miss the central combinatorial mechanism, optimize or prove the wrong target, or leave a crucial global obligation unclosed. 
In combination with the verifier results, this suggests that stronger Olympiad-level combinatorics systems need more than local algebraic or pattern-matching skill; they must improve target selection, invariant discovery, proof closure, and executable witness synthesis.



\section{Conclusion}

We present ComBench, a 100-problem Olympiad combinatorics benchmark that decouples \textit{Rigorous Proof Reasoning} from \textit{Constructive Realization} via a verifier-gated protocol combining rubric proof grading with deterministic witness checking. Frontier models exhibit a clear dissociation between the two capabilities---strong proofs can lack valid witnesses, while weaker proof grading can coexist with top construction performance---and below-credit proof failures concentrate on missing core mechanisms and wrong mathematical targets, calling for advances in global proof planning and explicit witness synthesis.


\section*{Limitations}
\label{sec:limitations}

ComBench is designed to make constructive combinatorial reasoning more directly evaluable, but it has several limitations. First, the deterministic verifier can only check a construction target that has been explicitly formalized in advance. This makes the evaluation reliable for the prescribed witness interface, but it does not cover every valid natural-language construction a model might describe. A model may have a mathematically meaningful construction idea that fails because it is not expressed in the required representation, while the verifier cannot judge alternative formulations outside its specification.

Second, ComBench focuses on Olympiad-level combinatorics. This focus is intentional, since combinatorics strongly stresses discrete structure, extremal reasoning, strategies, and explicit constructions. However, performance on ComBench should not be interpreted as a complete measure of Olympiad mathematics ability. Other domains such as algebra, geometry, and number theory require different forms of abstraction, symbolic manipulation, and proof technique, which are outside the scope of this benchmark.

\section*{Ethical Considerations}
\label{sec:ethical-considerations}

ComBench is built from public Olympiad-style combinatorics problems and does not contain personal, demographic, or sensitive user data. The main risks of this benchmark are therefore not privacy-related, but concern interpretation and use. Scores on ComBench should not be taken as a complete measure of mathematical intelligence, since the benchmark focuses on combinatorial reasoning and construction-centric tasks. In addition, some public problems may overlap with existing benchmarks or model training data, so we document overlap with IMO-Bench and encourage cautious interpretation of results on public-source problems.

All annotators involved in problem annotation, grading-guideline construction, construction specification, and verifier checking received compensation corresponding to their annotation workload.

The automatic evaluation pipeline also requires care. The LLM-based proof judge may disagree with expert human graders, and executable verifiers only check pre-specified construction targets rather than arbitrary natural-language mathematical reasoning. We therefore treat ComBench as a diagnostic benchmark for mathematical reasoning and witness realization, rather than as the sole basis for claims about model capability.

\bibliography{iclr2026_conference}

\appendix
\section{Dataset Details}
\label{app:dataset-statistics}

This appendix provides the aggregate dataset statistics supporting Section~\ref{sec:dataset}. We report the competition source distribution, year coverage, and category-by-type breakdown used in the main text. Full problem-level metadata, including the complete problem-to-category mapping, will be released with the public dataset rather than reproduced as a 100-row appendix table.

\begin{table}[h]
\centering
\small
\begin{tabular}{lr}
\toprule
Statistic & Count \\
\midrule
Total problems & 100 \\
Analysis-centric records & 50 \\
Construction-centric records & 50 \\
Primary categories & 5 \\
Competition sources & 15 \\
Year range & 2000--2025 \\
\bottomrule
\end{tabular}
\caption{Dataset summary statistics.}
\label{tab:app-dataset-summary}
\end{table}

\subsection{Source and Year Distribution}

Table~\ref{tab:app-source-distribution} reports the competition sources. The dataset combines canonical international problems with national and selection-test problems so that the benchmark is not dominated by a single contest family.

\begin{table}[h]
\centering
\small
\begin{tabular}{lr@{\qquad}lr}
\toprule
Source & Count & Source & Count \\
\midrule
IMO & 32 & CMO & 3 \\
TSTST & 14 & CGMO & 3 \\
USAMO & 13 & IMOSLC & 3 \\
USATST & 8 & USEMO & 2 \\
RMO & 6 & CTST & 2 \\
CSMO & 4 & ELMO & 2 \\
EGMO & 3 & EMC & 2 \\
APMO & 3 &  &  \\
\bottomrule
\end{tabular}
\caption{Competition source distribution in ComBench.}
\label{tab:app-source-distribution}
\end{table}

Table~\ref{tab:app-year-distribution} reports the year coverage. The distribution emphasizes recent Olympiad-style problems while retaining earlier problems that are useful for classical combinatorial themes.

\begin{table}[h]
\centering
\small
\begin{tabular}{lr}
\toprule
Year range & Count \\
\midrule
2000--2009 & 12 \\
2010--2014 & 11 \\
2015--2019 & 32 \\
2020--2023 & 33 \\
2024--2025 & 12 \\
\bottomrule
\end{tabular}
\caption{Year distribution in ComBench.}
\label{tab:app-year-distribution}
\end{table}

\subsection{Problem Type and Category Breakdown}

Table~\ref{tab:app-category-definitions} defines the five primary categories used in ComBench. Table~\ref{tab:app-category-type-breakdown} gives the category-by-type breakdown. The construction-centric portion is intentionally concentrated in categories where explicit witnesses naturally arise, especially extremal problems and existence-style tasks.

\begin{table}[h]
\centering
\small
\setlength{\tabcolsep}{3pt}
\begin{tabular}{p{0.30\columnwidth}p{0.58\columnwidth}}
\toprule
Category & Description \\
\midrule
Extremal Problems & Ask for maximum or minimum values under combinatorial constraints, often requiring both a sharp bound and a matching construction. \\
Existence and Construction & Ask whether an object exists or require an explicit witness satisfying a global condition. \\
Operations and Strategies & Involve games, processes, transformations, invariants, or strategic choices. \\
Graph Theory & Focus on graph structures, paths, colorings, connectivity, or discrete networks. \\
Counting & Require enumerative or combinatorial counting arguments, often with careful case distinctions or structural decompositions. \\
\bottomrule
\end{tabular}
\caption{Primary category definitions used in ComBench.}
\label{tab:app-category-definitions}
\end{table}

\begin{table}[h]
\centering
\small
\setlength{\tabcolsep}{3pt}
\begin{tabular}{lrrr}
\toprule
Category & Analysis & Construction & Total \\
\midrule
Extremal & 7 & 28 & 35 \\
Existence & 17 & 7 & 24 \\
Operations & 16 & 7 & 23 \\
Graph Theory & 6 & 5 & 11 \\
Counting & 4 & 3 & 7 \\
\midrule
Total & 50 & 50 & 100 \\
\bottomrule
\end{tabular}
\caption{Problem type and category breakdown. Analysis and Construction denote analysis-centric and construction-centric records.}
\label{tab:app-category-type-breakdown}
\end{table}

\subsection{Released Metadata Fields}

The public dataset release will include problem-level metadata and annotations needed to reproduce the aggregate analyses in this paper. Each released record will include the problem identifier, source, year, problem statement, problem type, primary category, reference answer, reference solution when available, and grading guidelines. Construction-centric records will additionally include the construction instruction, reference witness, verifier plan, and verifier code. These fields allow users to aggregate results by source, year, problem category, and evaluation target without requiring a long problem-level table in the paper appendix.

\section{Construction Annotation Pipeline}
\label{app:annotation-pipeline}

This appendix expands the three-stage construction-centric annotation pipeline summarized in Section~\ref{sec:dataset}. The pipeline converts a constructive combinatorics problem into a record containing a proof-grading rubric and an executable witness-verification interface.

\subsection{Stage 1: Specification and Rubric Construction}

Stage 1 starts from the original problem, reference answer, reference solution, human construction intent, and a reference witness. The construction intent specifies the type of object to be requested and the mathematical target that the object must realize. The reference witness provides an initial concrete construction before standardization.

\begin{table}[h]
\centering
\small
\setlength{\tabcolsep}{3pt}
\begin{tabular}{p{0.28\columnwidth}p{0.58\columnwidth}}
\toprule
Component & Role \\
\midrule
Problem and reference answer & Define the original mathematical target and final claim. \\
Reference solution & Supports rubric construction and proof-progress calibration. \\
Human intent & Specifies the desired witness type and construction target. \\
Reference witness & Provides a concrete construction to be standardized and checked. \\
\bottomrule
\end{tabular}
\caption{Stage 1 inputs.}
\label{tab:app-stage1-inputs}
\end{table}

Given these inputs, the LLM drafts three construction-side artifacts and one proof-side artifact: an item-specific construction instruction, a standardized reference construction, an informal verifier plan, and \(0/1/6/7\)-style grading guidelines. The instruction defines the expected witness and its representation. The standardized reference construction is the payload that should pass the final verifier. The verifier plan lists the format checks and mathematical constraints that the verifier should enforce. The grading guidelines define proof progress for the original problem.

\begin{table}[h]
\centering
\small
\setlength{\tabcolsep}{3pt}
\begin{tabular}{p{0.30\columnwidth}p{0.56\columnwidth}}
\toprule
Artifact & Purpose \\
\midrule
Instruction & Specifies what witness the model must output and in what format. \\
Reference construction & Provides a valid payload under the declared representation. \\
Verifier plan & Lists the intended format and mathematical checks. \\
Grading guidelines & Defines the \(0/1/6/7\)-point proof rubric. \\
\bottomrule
\end{tabular}
\caption{Stage 1 generated artifacts.}
\label{tab:app-stage1-artifacts}
\end{table}

Human review checks whether the construction instruction preserves the original problem meaning, whether the standardized reference construction satisfies the intended target, and whether the grading guidelines reflect meaningful Olympiad-style proof progress. Items that fail this review are returned to intent revision or Stage 1 generation.

\subsection{Stage 2: Verifier Generation and Semantic Audit}

Stage 2 generates deterministic Python verifier code from the reviewed problem statement, construction instruction, reference construction, and verifier plan. The verifier reads only the raw witness payload and checks both representation validity and mathematical validity. It should reject malformed payloads, incomplete objects, objects satisfying only local constraints, and objects that miss the intended global combinatorial condition.

After verifier generation, we run an LLM-assisted semantic audit over the full annotation chain. The audit compares the original problem, reference answer, construction intent, construction instruction, reference construction, verifier plan, and verifier code. Its goal is to determine whether the executable check still matches the same mathematical claim as the original problem.

\begin{table}[h]
\centering
\small
\setlength{\tabcolsep}{2pt}
\begin{tabular}{lp{0.52\columnwidth}}
\toprule
Audit target & Main check \\
\midrule
Construction target & Does it preserve the original mathematical claim? \\
Instruction & Is the requested witness precise and unambiguous? \\
Reference construction & Does it satisfy both the instruction and original target? \\
Verifier plan & Does it cover the necessary format and mathematical constraints? \\
Verifier code & Does it implement the intended checks without overfitting the reference construction? \\
\bottomrule
\end{tabular}
\caption{Semantic audit checklist.}
\label{tab:app-semantic-audit-checklist}
\end{table}

\begin{table}[h]
\centering
\small
\setlength{\tabcolsep}{3pt}
\begin{tabular}{lp{0.66\columnwidth}}
\toprule
Audit outcome & Meaning \\
\midrule
\textit{pass} & The construction target, instruction, witness, plan, and verifier are semantically aligned. \\
\textit{step1\_issue} & The construction intent, instruction, or reference construction changes, weakens, or distorts the original target. \\
\textit{step2\_issue} & The verifier misses constraints, checks the wrong property, conflicts with the instruction, or overfits the reference construction. \\
\bottomrule
\end{tabular}
\caption{Semantic audit outcomes.}
\label{tab:app-semantic-audit}
\end{table}

Records with \textit{step1\_issue} are returned to the specification stage, since the task definition itself is unreliable. Records with \textit{step2\_issue} are returned to verifier generation, since the construction target is acceptable but the executable checker is not.

\subsection{Stage 3: Assembly and Executable Reference Check}

Records passing semantic audit are assembled into the final JSONL format. Before acceptance, the standardized reference construction is passed to the verifier through the same raw-standard-input interface later used for model submissions. This executable reference check ensures that the stored construction, verifier code, and payload interface are mutually consistent.

The final human review focuses on verifier strictness, coverage, consistency, and diagnostic clarity. In particular, reviewers check that the verifier enforces global combinatorial constraints rather than merely validating syntax, rejects plausible but incomplete witnesses, and provides diagnostics that are useful for debugging the record. Additional verifier input and output details are provided in Appendix~\ref{app:verifier-details}.

\section{Record Schema and Quality Control}
\label{app:quality-control}

This appendix supports the record-design and quality-control claims in Section~\ref{sec:dataset}. ComBench records contain common fields for proof evaluation and additional fields for construction-centric evaluation.

\subsection{Record Schema}

\begin{table}[h]
\centering
\small
\setlength{\tabcolsep}{3pt}
\begin{tabular}{p{0.30\columnwidth}p{0.13\columnwidth}p{0.45\columnwidth}}
\toprule
Field & Applies to & Purpose \\
\midrule
\texttt{problem\_id}, source, year & All & Identify the original problem and support source/year analysis. \\
Problem statement & All & Provides the task given to the model. \\
Reference answer & All & Specifies the target answer or claim. \\
Reference solution & All & Supports rubric construction and proof judging when available. \\
Primary category & All & Supports category-level analysis. \\
Problem type & All & Indicates analysis-centric or construction-centric evaluation. \\
Grading guidelines & All & Defines the \(0/1/6/7\)-point proof rubric. \\
Construction intent & Constr. & Specifies the desired witness type and target. \\
Construction instruction & Constr. & Defines the model-facing witness request and output format. \\
Reference construction & Constr. & Provides the standardized witness used for executable reference checking. \\
Verifier plan & Constr. & Describes intended format and mathematical checks. \\
Verifier code & Constr. & Implements deterministic witness verification. \\
\bottomrule
\end{tabular}
\caption{ComBench record schema.}
\label{tab:app-record-schema}
\end{table}

The construction-centric fields define an executable interface rather than an additional natural-language hint. During evaluation, the model receives the construction instruction, submits one payload, and the verifier checks only that payload. The reference construction is used for dataset validation and is not provided to the response model.

\subsection{Example Construction-Centric Record}
\label{app:jsonl-example}

This subsection shows the complete information content of one construction-centric record. The released dataset stores these fields as JSONL; here they are formatted across paragraphs and code blocks for readability. The example is based on IMO 2020 Problem 4 with the fixed instance \(n=33\).

\begin{table}[h]
\centering
\small
\setlength{\tabcolsep}{3pt}
\begin{tabular}{p{0.30\columnwidth}p{0.56\columnwidth}}
\toprule
Field & Example value \\
\midrule
\texttt{id} & \texttt{IMO-2020-P4-33} \\
Source metadata & IMO 2020 Problem 4, instantiated with \(n=33\). \\
Problem type & Construction-centric. \\
Primary category & Graph Theory. \\
Reference answer & \(k=n^2-n+1\). \\
Verifier target & Check that both companies have 1056 valid monotone cable-car lines and no station pair is linked by both companies. \\
\bottomrule
\end{tabular}
\caption{Formatted metadata for an example construction-centric record.}
\label{tab:app-example-record-metadata}
\end{table}

\paragraph{Problem statement.}
There is an integer \(n > 1\). There are \(n^2\) stations on a slope of a mountain, all at different altitudes. Each of two cable car companies, A and B, operates \(k\) cable cars; each cable car provides a transfer from one of the stations to a higher one, with no intermediate stops. The \(k\) cable cars of A have \(k\) different starting points and \(k\) different finishing points, and a cable car which starts higher also finishes higher. The same conditions hold for B. We say that two stations are linked by a company if one can start from the lower station and reach the higher one by using one or more cars of that company, with no other movements between stations allowed. Determine the smallest positive integer \(k\) for which one can guarantee that there are two stations that are linked by both companies.

\paragraph{Reference answer.}
The reference answer is \(k=n^2-n+1\).

\paragraph{Construction instruction.}
For the specific case \(n=33\), determine the maximum number of cable cars \(k\) that each company can operate without creating any pair of stations linked by both companies. Provide an explicit construction for this maximum case. The \(33^2=1089\) stations are numbered 1 to 1089 from lowest to highest altitude. The output must be a single Python expression evaluating to a tuple of two sets, \texttt{(A\_lines, B\_lines)}. The sets \texttt{A\_lines} and \texttt{B\_lines} contain tuples \texttt{(i, j)}, representing a cable car line from station \(i\) to station \(j\) for companies A and B respectively. The configuration must satisfy all validity constraints for cable-car operations: distinct start points, distinct end points, and the monotonicity condition that a cable car starting higher also finishes higher. Each company must have exactly \(k=1056\) cars. The model must put only the Python expression inside the construction block, without markdown, explanations, or surrounding tags other than \texttt{<construct>...</construct>}.

\paragraph{Reference construction.} \mbox{}

\begin{lstlisting}[style=combpayload,backgroundcolor=\color{orange!8},rulecolor=\color{orange!55!black},caption={Reference construction payload for the example record.},label={lst:app-example-reference-construction}]
(
    {(i, i + 1) for i in range(1, 1090) if i % 33 != 0},
    {(i, i + 33) for i in range(1, 1057)}
)
\end{lstlisting}

\paragraph{Verifier code.}
The verifier reads the raw payload, parses it as a Python expression, checks the cable-car constraints for both companies, computes transitive closures, and rejects the construction if any linked pair appears for both companies.

\begin{lstlisting}[style=combcode,language=Python,caption={Complete verifier code for the example record.},label={lst:app-example-verifier-code}]
import sys

def verify():
    payload = sys.stdin.read().strip()
    if not payload:
        print("Empty input")
        return

    try:
        # Parse the payload directly as a Python expression.
        # eval is used because the reference construction uses comprehensions and range().
        data = eval(payload)
    except Exception as e:
        print(f"Failed to parse Python expression: {e}")
        return

    if type(data) is not tuple or len(data) != 2:
        print("Parsed object is not a tuple of exactly two elements")
        return

    A_lines, B_lines = data

    for name, lines in [("A_lines", A_lines), ("B_lines", B_lines)]:
        if type(lines) is not set:
            print(f"{name} is not a set")
            return
        for item in lines:
            if type(item) is not tuple or len(item) != 2:
                print(f"An item in {name} is not a tuple of exactly two elements")
                return
            u, v = item
            if type(u) is not int or type(v) is not int:
                print(f"An item in {name} contains non-integers")
                return

    if len(A_lines) != 1056:
        print(f"A_lines has {len(A_lines)} elements, expected 1056")
        return
    if len(B_lines) != 1056:
        print(f"B_lines has {len(B_lines)} elements, expected 1056")
        return

    for name, lines in [("A_lines", A_lines), ("B_lines", B_lines)]:
        u_set = set()
        v_set = set()
        sorted_lines = sorted(list(lines), key=lambda x: x[0])
        prev_v = -1

        for u, v in sorted_lines:
            if not (1 <= u <= 1089 and 1 <= v <= 1089):
                print(f"Station out of range in {name}: ({u}, {v})")
                return
            if u >= v:
                print(f"Invalid transfer in {name}: ({u}, {v}), must be u < v")
                return
            if u in u_set:
                print(f"Duplicate starting point {u} in {name}")
                return
            u_set.add(u)
            if v in v_set:
                print(f"Duplicate finishing point {v} in {name}")
                return
            v_set.add(v)
            if v <= prev_v:
                print(f"Monotonicity violated in {name}: sequence of v is not strictly increasing.")
                return
            prev_v = v

    def get_transitive_closure(lines):
        adj = {u: v for u, v in lines}
        closure = set()
        for start_node in adj:
            curr = start_node
            while curr in adj:
                nxt = adj[curr]
                closure.add((start_node, nxt))
                curr = nxt
        return closure

    A_closure = get_transitive_closure(A_lines)
    B_closure = get_transitive_closure(B_lines)

    intersection = A_closure.intersection(B_closure)
    if intersection:
        example = list(intersection)[0]
        print(f"Intersection is not empty. Found linked pairs in both companies, e.g., {example}.")
        return

    print("True")

if __name__ == "__main__":
    verify()
\end{lstlisting}

\paragraph{Grading guidelines.}
\textit{Partial credit} is assigned for the following forms of progress: (1) modeling each company's system as a directed graph on the stations and correctly deducing from the order-preserving and distinct start/endpoint conditions that each connected component is a path, possibly with isolated vertices; (2) deriving the component-count formula for one company, namely that with \(k\) edges on \(n^2\) vertices, the number of connected components is \(n^2-k\); and (3) giving the correct extremal pigeonhole step, using \(n^2=(n-1)(n+1)+1\) and \(n-1\) components to conclude that some component has at least \(n+1\) vertices.

\textit{Almost-correct credit} is assigned for solutions that prove sufficiency of \(k=n^2-n+1\) using the path-component structure and pigeonhole principle but make a small counting slip; or that present the full sufficiency proof and the idea of a counterexample for \(k=n^2-n\) but do not fully describe or verify the construction; or that are otherwise almost complete with non-negligible minor mistakes.

\paragraph{Reference solution.}
The answer is \(k=n^2-n+1\).

When \(k=n^2-n\), a valid construction exists and shows that this value is not sufficient. For \(n=33\), the reference construction above realizes this counterexample.

To prove that \(k=n^2-n+1\) is sufficient, view A and B as graphs whose vertices are the \(n^2\) stations and whose edges are the cable cars. For each company, the condition that a cable car starting higher also finishes higher, together with distinct starting points and distinct finishing points, implies that the connected components are paths, possibly with isolated vertices; edge direction is irrelevant for this component-counting argument.

If \(k=n^2-n+1\), then each company has exactly \(n^2-k=n-1\) connected components. Since \(n^2=(n-1)(n+1)+1\), the pigeonhole principle implies that some component of A has at least \(n+1\) vertices. Since B has only \(n-1\) connected components, this component of A must contain two vertices that belong to the same component of B. These two stations are therefore linked by both companies.

\subsection{Representative Witness Forms}
\label{app:witness-forms}

Construction-centric records use different payload styles depending on the mathematical object being requested. Table~\ref{tab:app-witness-form-summary} summarizes representative forms, and the following code blocks show the corresponding reference constructions.

\begin{table}[h]
\centering
\scriptsize  
\setlength{\tabcolsep}{1pt}
\begin{tabular}{
  >{\raggedright\arraybackslash\ttfamily}p{0.29\columnwidth}
  >{\raggedright\arraybackslash}p{0.33\columnwidth}
  >{\raggedright\arraybackslash}p{0.34\columnwidth}
}
\toprule
Record & Witness form & Payload meaning \\
\midrule
USAMO-2024-P4 & Explicit binary matrix & A fully enumerated 0/1 construction. \\
USAMO-2020-P4-10 & Ordered-pair list & A sequence of concrete ordered pairs. \\
TSTST-2024-P9-20 & Comprehension table & A large table represented by a compact rule. \\
IMOSLC-2023-P4-150 & Semantic dictionary & Structured fields with mathematical roles. \\
RMO-2019-P15 & Rule-based witness & Decision rule + supporting partitions. \\
\bottomrule
\end{tabular}
\caption{Representative witness payload forms in construction-centric records.}
\label{tab:app-witness-form-summary}
\end{table}

\paragraph{Explicit binary matrix.} \mbox{}
\begin{lstlisting}[style=combpayload,caption={Explicit binary-matrix witness from \texttt{USAMO-2024-P4}.},label={lst:app-witness-binary-matrix}]
[[0, 0, 0, 0],
 [0, 0, 0, 1],
 [0, 0, 1, 1],
 [0, 1, 1, 1],
 [1, 1, 1, 1]]
\end{lstlisting}

\paragraph{Explicit ordered-pair list.} \mbox{}
\begin{lstlisting}[style=combpayload,caption={Ordered-pair witness from \texttt{USAMO-2020-P4-10}.},label={lst:app-witness-pair-list}]
[[1, 0], [1, 1], [2, 1], [3, 1], [4, 1],
 [5, 1], [6, 1], [7, 1], [8, 1], [9, 1]]
\end{lstlisting}

\paragraph{Comprehension-defined table.} \mbox{}
\begin{lstlisting}[style=combpayload,caption={Comprehension-defined table from \texttt{TSTST-2024-P9-20}.},label={lst:app-witness-comprehension-table}]
[[((j - i) % 20) * 20 + i + 1 for j in range(20)]
 for i in range(20)]
\end{lstlisting}

\paragraph{Semantic dictionary.} \mbox{}
\begin{lstlisting}[style=combpayload,caption={Dictionary-structured witness from \texttt{IMOSLC-2023-P4-150}.},label={lst:app-witness-semantic-dict}]
{
  "endpoints": [150] + [
    val for i in range(2, 151)
    for val in ((i-1)*150 + i - 1, i*150)
  ],
  "rows": [[1]] + [[2*i - 1, 2*i - 2]
                   for i in range(2, 151)]
}
\end{lstlisting}

\paragraph{Rule-based witness.} \mbox{}
\begin{lstlisting}[style=combpayload,caption={Rule-based witness from \texttt{RMO-2019-P15}.},label={lst:app-witness-rule-based}]
{
  "rule": "lambda team: sum(1 for x in team if x in {1, 2, 3}) % 2 == 1",
  "partition_1": [
    [1, 2, 3, 4, 5, 6],
    [7, 8, 9, 10, 11, 12],
    [13, 14, 15, 16, 17, 18],
    [19, 20, 21, 22, 23, 24]
  ],
  "partition_3": [
    [1, 4, 5, 6, 7, 8],
    [2, 9, 10, 11, 12, 13],
    [3, 14, 15, 16, 17, 18],
    [19, 20, 21, 22, 23, 24]
  ]
}
\end{lstlisting}

\subsection{Quality-Control Checks}

\begin{table}[h]
\centering
\small
\setlength{\tabcolsep}{3pt}
\begin{tabular}{lp{0.62\columnwidth}}
\toprule
Check type & Required condition \\
\midrule
Target fidelity & The problem statement, reference answer, reference solution, grading guideline, construction instruction, and reference construction must target the same mathematical claim. \\
Verifier adequacy & The reference construction must be parseable and accepted by the verifier, and the verifier must enforce mathematical constraints rather than only format-level properties. \\
Dataset integrity & Source identifiers, problem types, categories, record counts, and final JSONL files must be mutually consistent. \\
\bottomrule
\end{tabular}
\caption{Quality-control checks.}
\label{tab:app-quality-control}
\end{table}

For construction-centric records, the most important risk is target drift: a construction task may accidentally ask for a weaker, stronger, or different object than the original problem requires. We therefore check the construction instruction and verifier against the original problem and reference answer, rather than only against the reference construction.

Verifier adequacy is checked at two levels. At the interface level, the verifier must accept the stored reference construction through the same raw-standard-input interface used for model submissions. At the mathematical level, it must enforce the global constraints specified by the construction instruction, not merely validate syntax or local consistency. Records that fail these checks are returned to the relevant stage of the annotation pipeline.

Dataset integrity checks ensure that the paper statistics match the evaluation files. We cross-check problem identifiers, source labels, year labels, problem categories, problem types, and construction-field availability against the final JSONL records used in the experiments.

\subsection{Human Audit Checklist}

The final human audit is applied after automatic generation and executable reference checking. Reviewers verify that the construction task preserves the original mathematical target, that the verifier checks the intended constraints rather than superficial syntax, and that the record metadata is consistent with the released dataset. Table~\ref{tab:app-human-audit-checklist} summarizes the questions used in this review.

\begin{table}[h]
\centering
\small
\setlength{\tabcolsep}{3pt}
\begin{tabular}{p{0.34\columnwidth}p{0.52\columnwidth}}
\toprule
Audit item & Question checked by reviewers \\
\midrule
Problem meaning & Does the construction task preserve the original mathematical claim? \\
Target strength & Is the requested witness neither weaker nor stronger than intended? \\
Output contract & Is the required representation explicit and unambiguous? \\
Reference construction & Does the stored construction satisfy the instruction and the original target? \\
Verifier coverage & Does the verifier check all essential mathematical constraints? \\
Verifier robustness & Does the verifier reject plausible invalid or incomplete witnesses? \\
Overfitting risk & Does the verifier check a class of valid objects rather than the stored reference only? \\
Diagnostics & Are failure messages sufficient for debugging annotation errors? \\
Metadata consistency & Do source, year, category, type, and construction fields match the released record? \\
\bottomrule
\end{tabular}
\caption{Human audit checklist.}
\label{tab:app-human-audit-checklist}
\end{table}

\section{Executable Verifier Details}
\label{app:verifier-details}
\label{app:construction-object-types}

This appendix describes the executable verifier used for construction-centric records. Here, a verifier refers to the deterministic Python code associated with an individual record. It is distinct from the rubric-guided proof judge: the proof judge evaluates natural-language reasoning, while the executable verifier checks only the submitted construction payload.

\subsection{Input and Output Contract}

For construction-centric records, the model response must contain exactly one payload inside a \texttt{<construct>...</construct>} block. The evaluation pipeline extracts the interior of this block and passes it as raw standard input to the item-specific Python verifier. The verifier does not read the proof text, the problem statement, or any surrounding explanation in the model response.

\begin{table}[h]
\centering
\small
\setlength{\tabcolsep}{3pt}
\begin{tabular}{p{0.28\columnwidth}p{0.22\columnwidth}p{0.38\columnwidth}}
\toprule
Component & Evaluator & Role \\
\midrule
Proof solution & Proof judge & Scores \textit{Rigorous Proof Reasoning}. \\
Construct \par payload & Code verifier & Checks \textit{Constructive Realization}. \\
Reference \par construction & Code verifier & Validates the record through executable reference checking. \\
\bottomrule
\end{tabular}
\caption{Separation between proof judging and executable construction verification.}
\label{tab:app-verifier-io}
\end{table}

The verifier returns a binary construction result. A construction passes if the verifier prints \texttt{True}; otherwise it fails. Missing, duplicated, or malformed \texttt{<construct>} blocks receive construction score \(0\) before verifier execution. During dataset validation, the stored reference construction is passed through the same raw-standard-input interface used for model submissions.

\subsection{Verifier Scope}

The executable verifier is not a general theorem prover and does not evaluate the natural-language proof. Its scope is limited to the construction target formalized by the item-specific instruction. Within this scope, the verifier checks three kinds of conditions.

\begin{table}[h]
\centering
\small
\setlength{\tabcolsep}{2pt}
\begin{tabular}{lp{0.52\columnwidth}}
\toprule
Responsibility & Typical checks \\
\midrule
Format validity & The payload can be parsed and has the declared type, dimensions, length, symbols, or value ranges. \\
Structural constraints & The submitted object satisfies representation-level constraints such as distinctness, legality of edges, valid coordinates, or complete assignments. \\
Mathematical correctness & The object satisfies the global combinatorial properties required by the construction instruction, such as coverage, extremality, non-conflict, invariance, or counterexample validity. \\
\bottomrule
\end{tabular}
\caption{Verifier responsibilities.}
\label{tab:app-verifier-responsibilities}
\end{table}

This design intentionally separates proof quality from witness validity. A model may receive partial proof credit while failing the verifier, and a valid witness does not by itself guarantee a complete proof. The combined score for construction-centric records is defined in Section~\ref{sec:score-integration}.

\subsection{Construction Object Types}

Construction-centric records use several recurring witness representations. Table~\ref{tab:app-construction-object-types} summarizes representative object types and the corresponding verifier checks.

\begin{table}[h]
\centering
\small
\setlength{\tabcolsep}{2pt}
\begin{tabular}{lp{0.52\columnwidth}}
\toprule
Object type & Typical verifier checks \\
\midrule
Matrices or grids & Shape, allowed entries, row/column constraints, coverage, adjacency, or path-count constraints. \\
Colorings & Domain coverage, color validity, forbidden monochromatic patterns, adjacency constraints, or extremal requirements. \\
Graphs & Vertex and edge validity, degree constraints, connectivity, subgraph avoidance, paths, cycles, or coloring properties. \\
Selected cell sets & Coordinate validity, distinctness, board constraints, coverage, domination, or packing properties. \\
Permutations or sequences & Length, element set, uniqueness, recurrence conditions, monotonicity, or extremal objective values. \\
Set systems & Membership validity, intersection constraints, covering properties, and cardinality requirements. \\
Strategies & Legal move sequence, game-state transition validity, terminal condition, and claimed guarantee. \\
Counterexamples & Format validity and direct falsification of the target statement under the specified parameters. \\
\bottomrule
\end{tabular}
\caption{Representative construction object types.}
\label{tab:app-construction-object-types}
\end{table}

\section{Evaluation Protocol}
\label{app:evaluation-details}

This appendix expands the evaluation configuration in Section~\ref{sec:experiments}. It specifies the response-model sampling setup, prompt templates, parsing rules, failure handling, and metric definitions used to reproduce the reported results.

\subsection{Response Models and Sampling}

The response models are the models reported in Table~\ref{tab:main-results}. For each record, we sample four independent solutions using temperature \(0.6\) and model-specific maximum token limits. All response models are evaluated with the same downstream protocol.

Proof evaluation uses a unified rubric-guided Gemini-3.1-Pro-Preview-Thinking judge for all response models. The judge receives the problem statement, the reference solution when available, and the problem-specific grading guidelines. Construction evaluation uses the item-specific deterministic Python verifier associated with each construction-centric record.

\subsection{Prompt Templates}
\label{app:prompt-templates}

The prompt format depends on the record type. The following templates summarize the exact structure used in evaluation; bracketed fields are filled with the corresponding record-specific content.

\paragraph{Analysis-centric records.}\mbox{}
\begin{lstlisting}[style=combcode]
You are solving one IMO-style math problem.

## Problem
"""
Please answer the problem adhering to the following rules:
1. Please use LaTeX format to represent the variables and formulas used in the solution process and results.
2. Please provide a complete and explicit solution process in the response.
3. In the end of the response:
    - If the problem has final answer(s), put them in \boxed{}.
    - If the problem requires multiple answers, list them in order, each in a separate \boxed{}.
    - If the problem is a proof or does not require a final numerical answer, provide a complete and rigorous proof, and do not use \boxed{}.

Problem:
[Original problem statement]
"""

You must answer using exactly the following output structure:

## Solution
[Your complete solution, including the final answer if applicable.]

Rules:
- Output the solution heading exactly as written above.
- Do not use `## Solution to Question 1` or `## Solution to Question 2`.
- Do not use any <construct>...</construct> block.
\end{lstlisting}

\paragraph{Construction-centric records.}\mbox{}

\begin{lstlisting}[style=combcode]
You are solving one IMO-style math problem with an additional construction deliverable.

## Question 1
"""
[Original problem statement]
"""

## Question 2
"""
[Item-specific construction instruction]
"""

You must answer using exactly the following output structure:

## Solution to Question 1
[Your proof and final answer to Question 1 only.]

## Solution to Question 2
[Your response to Question 2. This section must include exactly one construction block.]
<construct>
[Construction payload]
</construct>

Rules:
- Output the two solution headings exactly as written above.
- Do not put any <construct> block in Solution to Question 1.
- Put exactly one <construct>...</construct> block in Solution to Question 2.
- Inside <construct>...</construct>, put ONLY the construction payload and nothing else.
- Do not use code fences.
\end{lstlisting}

The proof part evaluates \textit{Rigorous Proof Reasoning}. The construction payload evaluates \textit{Constructive Realization}.

\paragraph{Rubric-guided proof evaluation.}
All response models are graded with the same proof-evaluation template. The judge receives the problem statement, the ground-truth solution when available, the item-specific grading guidelines, and the proposed solution. It must return exactly one parsed score from the four-level rubric:
\begin{lstlisting}[style=combcode]
You are an expert grader for the International Mathematics Olympiad (IMO).
Evaluate the proposed solution strictly and rigorously.

Inputs:
- Problem Statement: [Original problem statement]
- Ground-Truth Solution: [Reference solution, if available]
- Specific Grading Guidelines: [Problem-specific 0/1/6/7 rubric]
- Proposed Solution: [Model response]

Output the final score using exactly one of:
<points>7 out of 7</points>
<points>6 out of 7</points>
<points>1 out of 7</points>
<points>0 out of 7</points>
\end{lstlisting}

We do not use final-answer equivalence grading as the main proof score; the reported proof scores use the rubric-guided proof judge.

\subsection{Parsing and Failure Handling}

For analysis-centric records, the parser extracts the model's proof solution and sends it to the proof judge. For construction-centric records, the parser extracts the proof section and the unique \texttt{<construct>...</construct>} block. Only the block interior is passed to the executable verifier.

If a construction-centric response has no construction block, multiple construction blocks, or a malformed block, its construction score is \(0\) without verifier execution. The proof text is still judged when it can be parsed. Missing or non-parseable outputs remain in the denominator for aggregate metrics. For SU-01, only outputs satisfying the evaluation format are treated as valid parsed outputs; missing or non-parseable outputs are counted under the same full-denominator accounting.

\subsection{Scoring and Metrics}

The proof judge returns one of \(0,1,6,7\) points out of \(7\). The construction verifier returns a binary result: \(1\) if the submitted payload is accepted and \(0\) otherwise. For construction-centric records, the aggregate score uses the verifier-gated rule defined in Section~\ref{sec:score-integration}.

The main text reports two metrics. \textit{Avg.} is the average normalized score over all generated solutions. \textit{Best@4} first takes the highest score among four sampled solutions for each record and then averages these per-record best scores. Additional reports may include \textit{Pass@4}, the percentage of records solved in at least one of four sampled solutions; \textit{Pass\textasciicircum{}4}, the percentage of records solved in all four sampled solutions; and construction pass rate, the per-sample percentage of construction payloads accepted by the verifier.

\section{Additional Experimental Results}
\label{app:additional-results}

This appendix provides supplementary results for the main-result and subset-analysis claims in Section~\ref{sec:experiments}. The main text reports the most compact tables and figures; here we include additional category-level, problem-level, and repeated-sampling summaries.

\subsection{Category-Level Summary}

Table~\ref{tab:app-category-summary} summarizes average category performance across the evaluated models. The values use the same scoring rule as Table~\ref{tab:main-results}: analysis-centric records use proof scores, while construction-centric records use verifier-gated scores.

\begin{table}[h]
\centering
\small
\resizebox{\columnwidth}{!}{%
\begin{tabular}{lrrlr}
\toprule
Category & \# records & Mean Avg. & Best model & Best Avg. \\
\midrule
Extremal Problems & 35 & 34.1 & GPT-5.5 & 69.5 \\
Existence and Construction & 24 & 22.9 & GPT-5.5 & 48.1 \\
Operations and Strategies & 23 & 31.6 & GPT-5.5 & 65.5 \\
Graph Theory & 11 & 59.6 & Gemini-3.1-Pro & 75.3 \\
Counting & 7 & 59.4 & GPT-5.5 & 94.4 \\
\bottomrule
\end{tabular}
}
\caption{Supplementary category-level performance summary.}
\label{tab:app-category-summary}
\end{table}

The category-level averages reinforce the trend shown in Figure~\ref{fig:category-performance-top4}. \textit{Existence and Construction} is the lowest-scoring category on average, even though it is not the largest category. In contrast, \textit{Counting} and \textit{Graph Theory} have substantially higher average scores, suggesting that current models are more reliable on locally structured enumeration or graph-structural arguments than on global existence and witness-realization problems.

\subsection{Problem-Level Difficulty}

Problem-level success counts are computed over four sampled solutions per model. For analysis-centric records, a sampled solution is counted as solved when the proof score is \(7\). For construction-centric records, strict success requires both proof score \(7\) and construction score \(1\).

For analysis-centric records, the easiest problems by total full-proof samples include \texttt{IMO2008\_5} and \texttt{TSTST2023\_4}, each with 32 full-score samples across the evaluated models. The hardest analysis-centric problems include \texttt{IMO2024\_5}, \texttt{IMO2018\_3}, \texttt{IMO2017\_3}, \texttt{IMO2013\_2}, \texttt{IMO2012\_3}, \texttt{IMO2009\_6}, \texttt{USAMO2025\_6}, and \texttt{TSTST2022\_5}, each with zero full-score samples.

For construction-centric records, the easiest strict problems include \texttt{RMO-2019-P15} with 31 strict successful samples, \texttt{APMO-2012-P2} and \texttt{CSMO-2018-P14-789} with 27, and \texttt{USAMO-2000-P4} and \texttt{CSMO-2004-P7} with 25. The hardest strict construction-centric problems include \texttt{IMO-2025-P6-23}, \texttt{IMO-2022-P6}, \texttt{IMO-2004-P3}, \texttt{USAMO-2022-P6-202}, \texttt{USAMO-2009-P3-20-21}, \texttt{CTST-2022-P7-518}, \texttt{RMO-2018-P4-81}, and \texttt{RMO-2009-P7}, each with zero strict successful samples.

\subsection{Construction-Centric Stability}

The construction-centric subset shows a large gap between one-success and all-success repeated-sampling metrics. For example, GPT-5.5 reaches \(68.0\%\) verifier-gated Pass@4 but only \(42.0\%\) Pass\textasciicircum{}4. This indicates that many construction-centric records can be solved by at least one sampled solution, but are not solved consistently across all four samples. The gap supports the use of both average scores and best-of-four scores in the main text: the former measures typical performance, while the latter measures whether the model can find a correct solution under repeated sampling.

\section{Proof Error Taxonomy}
\label{app:error-taxonomy}

This appendix supports the error analysis in Section~\ref{sec:experiments}. The taxonomy applies only to proof-side critiques, and therefore diagnoses failures in \textit{Rigorous Proof Reasoning}. Failures of \textit{Constructive Realization} are analyzed separately through construction pass rates, verifier-gated scores, and the case studies in Appendix~\ref{app:case-study}.

\subsection{Scope and Annotation Rule}

We analyze every below-full-credit proof sample, including proofs scored \(0\), \(1\), or \(6\), as well as proof components that are unscored because of parsing or evaluation failure. Each sample is assigned exactly one primary reason. Primary reasons are mutually exclusive by construction, while secondary tags may overlap. For the paper-facing distribution, mathematical categories below the global \(2\%\) frequency threshold are merged into \textit{Other Mathematical Error}. \textit{Format or Unscored} is kept separate because it is not a mathematical proof error.

\subsection{Taxonomy Definitions}

Table~\ref{tab:app-error-taxonomy} defines the proof-error categories. The examples are schematic: they illustrate the kind of mathematical failure covered by each category, rather than quoting a particular model response.

\begin{table}[h]
\centering
\scriptsize
\setlength{\tabcolsep}{3pt}
\begin{tabular}{p{0.31\columnwidth}p{0.34\columnwidth}p{0.25\columnwidth}}
\toprule
Error type & Meaning & Example pattern \\
\midrule
Missing Core Mechanism & The solution misses the central combinatorial idea, such as an invariant, extremal choice, bijection, or construction principle. & Gives local observations but never identifies the invariant needed to control the process. \\
Wrong Mathematical Target & The proof solves a different statement, optimizes the wrong quantity, or uses an incorrect final value. & Proves a lower bound for \(n-1\) when the problem asks for the exact maximum. \\
False Lemma & The argument relies on a false claim or an unsupported theorem-like assertion. & Claims that every maximal configuration is automatically balanced, without proof and in fact falsely. \\
Faulty Induction & The induction base case, transition, preserved invariant, or parameter is incorrect. & Assumes the induction step after deleting an object, but the remaining structure no longer satisfies the hypothesis. \\
Unjustified Leap & A crucial implication is asserted without enough mathematical justification. & States that a greedy choice can always be extended, but gives no exchange or obstruction argument. \\
Incomplete Case Analysis & Necessary cases are omitted or treated asymmetrically. & Handles one parity case and concludes the result for all parities. \\
Proof Closure Gap & The response has meaningful progress but fails to close the final implication, optimality proof, or equality case. & Proves an upper bound but never supplies the matching construction. \\
Insufficient Mathematical Progress & The response contains only superficial reformulation or weak observations. & Restates the problem and checks small cases without a route to a proof. \\
Format or Unscored & The proof component cannot be scored because of missing content, parsing failure, or evaluator failure. & The required solution section is absent or non-parseable. \\
Other Mathematical Error & Low-frequency mathematical errors not covered by the categories above. & A specialized counting or notation error that does not fit the main categories. \\
\bottomrule
\end{tabular}
\caption{Proof-error taxonomy used for below-full-credit proof samples.}
\label{tab:app-error-taxonomy}
\end{table}

\subsection{Global Distribution}

Table~\ref{tab:app-error-distribution} reports the primary reason distribution after merging low-frequency mathematical categories into \textit{Other Mathematical Error}. The largest category is \textit{Missing Core Mechanism}, followed by \textit{Wrong Mathematical Target}. This pattern indicates that most failures are substantive mathematical failures rather than formatting failures.

\begin{table}[h]
\centering
\small
\resizebox{0.50\columnwidth}{!}{%
\begin{tabular}{lrr}
\toprule
Primary reason & Count & Share \\
\midrule
Missing Core Mechanism & 1213 & 41.2 \\
Wrong Mathematical Target & 588 & 20.0 \\
Format or Unscored & 324 & 11.0 \\
False Lemma & 290 & 9.8 \\
Faulty Induction & 202 & 6.9 \\
Unjustified Leap & 144 & 4.9 \\
Incomplete Case Analysis & 114 & 3.9 \\
Other Mathematical Error & 70 & 2.4 \\
\bottomrule
\end{tabular}
}
\caption{Primary proof-error distribution. Shares are percentages.}
\label{tab:app-error-distribution}
\end{table}

\subsection{Model-Level Error Profiles}

Table~\ref{tab:app-error-model-rate} reports how many below-full-credit proof samples each model produces. Table~\ref{tab:app-error-model-counts} further breaks these samples down by primary reason. The counts show that stronger models still fail mainly by missing the core mechanism or targeting the wrong mathematical claim, while models with many format or parsing failures have a different error profile.

\begin{table}[h]
\centering
\small
\resizebox{0.50\columnwidth}{!}{%
\begin{tabular}{lrr}
\toprule
Model & Below-full samples & Rate \\
\midrule
GPT-5.5 & 174/400 & 43.5 \\
Gemini-3.1-Pro & 203/400 & 50.8 \\
Kimi-K2.6 & 245/400 & 61.2 \\
DeepSeek-V4-Pro & 271/400 & 67.8 \\
SU-01 & 319/400 & 79.8 \\
Qwen3.6-Max & 333/400 & 83.2 \\
GLM-5.1 & 336/400 & 84.0 \\
Nemotron & 349/400 & 87.2 \\
Qwen3.6-35B & 352/400 & 88.0 \\
Gemma-4-31B-IT & 363/400 & 90.8 \\
\bottomrule
\end{tabular}
}
\caption{Below-full-credit proof samples by model.}
\label{tab:app-error-model-rate}
\end{table}

\begin{table}[h]
\centering
\scriptsize
\setlength{\tabcolsep}{2pt}
\resizebox{0.50\columnwidth}{!}{%
\begin{tabular}{lrrrrrrrr}
\toprule
Model & MCM & WMT & FL & FI & UL & ICA & Fmt & Other \\
\midrule
GPT-5.5 & 85 & 32 & 23 & 14 & 8 & 9 & 0 & 3 \\
Gemini-3.1-Pro & 120 & 29 & 17 & 8 & 13 & 9 & 4 & 3 \\
Kimi-K2.6 & 108 & 44 & 43 & 16 & 10 & 16 & 5 & 3 \\
DeepSeek-V4-Pro & 107 & 42 & 37 & 41 & 10 & 18 & 11 & 5 \\
SU-01 & 49 & 31 & 21 & 19 & 3 & 7 & 187 & 2 \\
Qwen3.6-Max & 139 & 58 & 34 & 26 & 18 & 13 & 38 & 7 \\
GLM-5.1 & 143 & 88 & 30 & 17 & 13 & 11 & 26 & 8 \\
Nemotron & 122 & 106 & 52 & 26 & 22 & 12 & 4 & 5 \\
Qwen3.6-35B & 160 & 83 & 14 & 20 & 24 & 11 & 28 & 12 \\
Gemma-4-31B-IT & 180 & 75 & 19 & 15 & 23 & 8 & 21 & 22 \\
\bottomrule
\end{tabular}
}
\caption{Model-level primary proof-error counts. Abbreviations: MCM = Missing Core Mechanism; WMT = Wrong Mathematical Target; FL = False Lemma; FI = Faulty Induction; UL = Unjustified Leap; ICA = Incomplete Case Analysis; Fmt = Format or Unscored.}
\label{tab:app-error-model-counts}
\end{table}

\section{Overlap with IMO-Bench}
\label{app:imobench-overlap}

This appendix documents the overlap analysis between ComBench and IMO-Bench. The purpose is benchmark-coverage transparency: both benchmarks draw from public Olympiad-style sources, so we report how much of ComBench overlaps with IMO-Bench and how the overlap was determined.

\subsection{Matching Protocol}

We compare all 100 ComBench problems against IMO-Bench using three complementary signals. First, we check source metadata, including competition name, year, and problem number when available. Second, we run problem-statement similarity search to identify candidates whose wording is close even when source identifiers differ. Third, we manually inspect high-similarity candidates to distinguish true overlap from problems that share only common combinatorial terminology.

A ComBench problem is counted as overlapping if it is an exact source match or a substantially modified version of an IMO-Bench item. Problems with similar themes but different mathematical targets are not counted as overlapping.

\subsection{Overlap Summary}

\begin{table}[h]
\centering
\small
\setlength{\tabcolsep}{4pt}
\begin{tabular}{lr}
\toprule
Judgment category & Count \\
\midrule
Source-based duplicate & 6 \\
Text-similarity duplicate & 6 \\
Manual-review duplicate & 2 \\
Inspected but not duplicate & 4 \\
No high-confidence match & 82 \\
\midrule
Total & 100 \\
\bottomrule
\end{tabular}
\caption{Overlap judgments against IMO-Bench.}
\label{tab:app-imobench-overlap-summary}
\end{table}

The final overlap rate is \(14/100\). The remaining 86 problems have no high-confidence match after source-based matching, text-similarity search, and manual inspection of ambiguous candidates. This indicates that ComBench provides largely complementary coverage to IMO-Bench while documenting the overlapping subset at the metadata level.

\subsection{Released Metadata and Filtering}

We do not reproduce the full matched-problem list in the appendix. Instead, the released metadata will include, for each inspected ComBench problem, the best-matching IMO-Bench entry when applicable, the match type, the similarity score, and the manual decision note. This allows users to filter or analyze overlapping records without using appendix space for a problem-level listing.

The overlap analysis is used to characterize benchmark coverage rather than to exclude records. Overlapping source problems are kept because ComBench adds construction-centric instructions, reference witnesses, and deterministic verifier-gated scoring; these record-level evaluation targets may differ from those in IMO-Bench even when the source problem overlaps.

\section{Manual Audit of Automatic Proof Evaluation}
\label{app:human-auto-alignment}

This appendix describes the sampled manual audit used to check the reliability of the automatic proof judge. The goal is not to replace expert mathematical review, but to verify that the rubric-guided proof judge produces reasonable scores on a representative subset of model outputs.

\subsection{Audit Sampling Protocol}

We sample proof-evaluation results from the 10 response models used in the main experiments. For each model, we randomly select five proof samples from the set of scored proof evaluations, excluding outputs whose proof score is null or unscored because of parsing or evaluation failure. This yields 50 audited proof samples in total.

The sampling covers both analysis-centric and construction-centric records. The audit only checks the proof-judge decision: whether the assigned proof score is consistent with the problem statement, reference solution when available, and problem-specific grading guidelines. Construction-verifier decisions are not included in this audit, because they are deterministic executable checks and are analyzed separately through construction pass rates and verifier-gated scores.

\subsection{Audit Decision Labels}

Each sampled proof evaluation is assigned one of three manual-audit labels:

\begin{itemize}
    \item \textbf{Accepted}: the automatic proof score is consistent with the rubric and the judge rationale is mathematically reasonable.
    \item \textbf{Minor disagreement}: the score is debatable near a rubric boundary, but the decision does not change the qualitative interpretation of the result.
    \item \textbf{Rejected}: the automatic proof score is clearly inconsistent with the rubric or misses a major mathematical issue.
\end{itemize}

We report the strict acceptance rate as the fraction of audited samples labeled \textit{Accepted}. We also report an accepted-or-minor rate, which treats boundary disagreements as acceptable for aggregate trend analysis.

\subsection{Audit Summary}

Table~\ref{tab:app-manual-audit-summary} summarizes the audit results. The released audit metadata contains the sampled model output, the proof-judge rationale, the assigned proof score, and the manual-audit decision for each sampled case. In this sample, 45 of 50 automatic proof-judge decisions are accepted by the manual audit. Four additional cases are marked as minor boundary disagreements, and one case is rejected.

\begin{table}[h]
\centering
\small
\resizebox{0.80\columnwidth}{!}{%
\begin{tabular}{lrrrrr}
\toprule
Model & Samples & Accepted & Minor & Rejected & Accept rate (\%) \\
\midrule
GPT-5.5 & 5 & 4 & 1 & 0 & 80.0 \\
Gemini-3.1-Pro & 5 & 4 & 1 & 0 & 80.0 \\
Kimi-K2.6 & 5 & 3 & 1 & 1 & 60.0 \\
DeepSeek-V4-Pro & 5 & 5 & 0 & 0 & 100.0 \\
SU-01 & 5 & 4 & 1 & 0 & 80.0 \\
Qwen3.6-Max & 5 & 5 & 0 & 0 & 100.0 \\
GLM-5.1 & 5 & 5 & 0 & 0 & 100.0 \\
Nemotron & 5 & 5 & 0 & 0 & 100.0 \\
Qwen3.6-35B & 5 & 5 & 0 & 0 & 100.0 \\
Gemma-4-31B-IT & 5 & 5 & 0 & 0 & 100.0 \\
\midrule
Total & 50 & 45 & 4 & 1 & 90.0 \\
\bottomrule
\end{tabular}
}
\caption{Manual audit summary for sampled automatic proof evaluations.}
\label{tab:app-manual-audit-summary}
\end{table}

This audit is intended as a reliability check for large-scale automatic proof grading. The strict acceptance rate is \(90.0\%\), and the accepted-or-minor rate is \(98.0\%\). These results support the use of the automatic judge for aggregate benchmarking, while also showing that boundary disagreements and occasional grading errors remain possible. The audit is limited to sampled proof evaluations and should not be interpreted as a guarantee that the automatic judge is error-free on every record.

\section{Case Studies for Verifier-Gated Scoring}
\label{app:case-study}

This appendix gives representative construction-centric case studies illustrating why verifier-gated scoring is used. The cases cover different proof scores, construction outcomes, and response models.

\subsection{Case 1: Proof Score 6 and Construction Score 0}

\begin{table}[h]
\centering
\small
\setlength{\tabcolsep}{3pt}
\begin{tabular}{lp{0.56\columnwidth}}
\toprule
Item & Description \\
\midrule
Problem & \texttt{IMO-2022-P6} \\
Run & GPT-5.5, run 1 \\
Proof score & \(6/7\) \\
Construction result & \(0/1\) \\
Verifier-gated score & \(1/7\) \\
\bottomrule
\end{tabular}
\caption{Case study with proof score 6 and failed construction.}
\label{tab:app-case-study-proof6}
\end{table}

\begin{mdframed}[style=mdpurplebox,frametitle={Problem IMO-2022-P6}]
Let \(n\) be a positive integer. A Nordic square is an \(n \times n\) board containing all the integers from \(1\) to \(n^2\) so that each cell contains exactly one number. An uphill path is a sequence of one or more cells such that:
\begin{enumerate}[leftmargin=*]
    \item the first cell in the sequence is a valley, meaning the number written is less than all its orthogonal neighbors;
    \item each subsequent cell in the sequence is orthogonally adjacent to the previous cell; and
    \item the numbers written in the cells in the sequence are in increasing order.
\end{enumerate}
Find, as a function of \(n\), the smallest possible total number of uphill paths in a Nordic square.
\end{mdframed}

\paragraph{Relevant grading guidelines.}
\begin{itemize}[leftmargin=*]
    \item \textit{Partial:} Establish the lower-bound idea by associating adjacent pairs with uphill paths, correctly count the \(2n(n-1)\) adjacent pairs in the grid, and add at least one length-one uphill path to obtain \(2n^2-2n+1\).
    \item \textit{Almost:} Give the full lower bound \(2n^2-2n+1\) and an intended matching construction using a set or tree structure, but leave a gap in justifying that the required construction exists for all \(n\), or leave a gap in the cropping/adaptation argument.
\end{itemize}

\paragraph{Full GPT-5.5 generation.}\mbox{}

\textbf{Solution to Question 1.}
Let the \(n^2\) cells of the board be regarded as the vertices of the usual \(n\times n\) grid graph \(G_n\). Thus two vertices are adjacent if and only if the corresponding cells are orthogonally adjacent. The number of edges of \(G_n\) is
\[
|E(G_n)|=n(n-1)+n(n-1)=2n(n-1).
\]
Put an orientation on every edge from the smaller number to the larger number. Since all numbers are distinct, this gives an acyclic orientation of \(G_n\). A valley is precisely a source of this orientation.

For a cell \(v\), let \(p(v)\) be the number of uphill paths ending at \(v\). If \(v\) is a valley, then \(p(v)=1\), corresponding to the path consisting only of \(v\). If \(v\) is not a valley, then every uphill path ending at \(v\) is obtained uniquely by taking an uphill path ending at a smaller adjacent cell and then appending \(v\). Hence
\[
p(v)=\sum_{u\to v} p(u),
\]
where the sum is over all directed edges entering \(v\).

The total number of uphill paths is therefore
\[
P=\sum_v p(v).
\]
Since every cell is reachable by following decreasing numbers until one reaches a valley, we have \(p(v)\ge 1\) for every \(v\). Moreover, every directed edge \(u\to v\) contributes at least one uphill path ending at \(v\), namely any uphill path ending at \(u\) extended by the edge \(u\to v\). Thus
\[
P\ge 1+|E(G_n)|=1+2n(n-1).
\]
It remains to show that this lower bound is attainable.

Choose an independent set \(S\) of cells such that \(G_n-S\) is a tree. Such a set exists for every rectangular grid; one may construct it by the standard zig-zag deletion pattern, deleting one cell from each elementary square in such a way that no two deleted cells are adjacent and the remaining graph is connected and acyclic.

Orient the tree \(G_n-S\) away from one chosen root cell \(r\). Then every cell of \(G_n-S\) except \(r\) has exactly one incoming edge inside the tree. Finally orient every edge incident with a cell of \(S\) toward that cell. Because \(S\) is independent, no edge joins two cells of \(S\). Thus every cell of \(S\) is a sink.

Now assign the numbers \(1,2,\dots,n^2\) in any order compatible with this acyclic orientation. The unique valley is the root \(r\). Every cell of \(G_n-S\) has exactly one smaller neighbor, namely its parent in the rooted tree, while every cell of \(S\) has only larger incoming contributions from cells with exactly one uphill path reaching them.

Therefore every edge contributes exactly one new uphill path, and together with the trivial path at the unique valley, the total number of uphill paths is
\[
1+|E(G_n)|=1+2n(n-1).
\]
Hence the smallest possible total number of uphill paths is
\[
\boxed{2n(n-1)+1}.
\]

\paragraph{Scoring interpretation.}
The proof contains the main lower-bound argument and an intended construction mechanism, so the proof-side score of \(6/7\) is consistent with the almost-complete guideline. However, the constructive part of the proof remains high-level: it asserts a ``standard zig-zag deletion pattern'' rather than fully specifying and verifying the required object. The construction-centric response must provide a complete witness in the prescribed representation, and it does not receive construction credit. The verifier-gated demotion from \(6\) to \(1\) is therefore appropriate: the response shows substantial \textit{Rigorous Proof Reasoning}, but it does not realize the construction as a complete, mechanically checkable object.

\subsection{Case 2: Proof Score 7 and Construction Score 0}

\begin{table}[h]
\centering
\small
\setlength{\tabcolsep}{3pt}
\begin{tabular}{lp{0.56\columnwidth}}
\toprule
Item & Description \\
\midrule
Problem & \texttt{USAMO-2009-P3-20-21} \\
Run & GPT-5.5, run 1 \\
Proof score & \(7/7\) \\
Construction result & \(0/1\) \\
Verifier-gated score & \(6/7\) \\
\bottomrule
\end{tabular}
\caption{Case study with proof score 7 and failed construction.}
\label{tab:app-case-study-proof7}
\end{table}

\begin{mdframed}[style=mdpurplebox,frametitle={Problem USAMO-2009-P3}]
We define a \emph{chessboard polygon} to be a simple polygon whose sides are situated along lines of the form \(x=a\) or \(y=b\), where \(a\) and \(b\) are integers. These lines divide the interior into unit squares, which are shaded alternately gray and white so that adjacent squares have different colors. To tile a chessboard polygon by dominoes is to exactly cover the polygon by non-overlapping \(1 \times 2\) rectangles. Finally, a \emph{tasteful tiling} is one which avoids the two forbidden configurations of dominoes and colors. 

Prove that (a) if a chessboard polygon can be tiled by dominoes, then it can be done so tastefully, and (b) such a tasteful tiling is unique.
\end{mdframed}

\paragraph{Relevant grading guidelines.}
\begin{itemize}[leftmargin=*]
    \item \textit{Partial:} For existence, use the lower-left-square idea and justify the key local switch in a \(2\times2\) block. For uniqueness, overlay two tilings and reduce the argument to ruling out alternating cycles under tastefulness.
    \item \textit{Almost:} Give an essentially complete induction proof of existence with a local gap, or prove uniqueness by an alternating-loop method while leaving a localized topological gap.
    \item \textit{Full:} A complete proof of existence and uniqueness without such gaps receives full proof credit.
\end{itemize}

\paragraph{Full GPT-5.5 generation.}\mbox{}

\textbf{Solution to Question 1.}
We prove the result by using the standard height-function method for domino tilings.

Let \(P\) be a chessboard polygon which is tileable by dominoes. Since \(P\) is a simple polygon, the union of its unit squares is simply connected.

Color the unit squares alternately white and black. For every edge of the square grid contained in \(P\), orient it so that a white square lies on its left, equivalently a black square lies on its right. For a boundary edge, this convention uses the unique square of \(P\) adjacent to that edge.

Given a domino tiling \(T\), define a height function
\[
h_T : \{\text{lattice vertices of }P\} \to \mathbb Z
\]
as follows. Fix one boundary vertex \(v_0\) and set
\[
h_T(v_0)=0.
\]
If \(e=uv\) is an oriented grid edge from \(u\) to \(v\), write \(e\in T\) when \(e\) is crossed by a domino of \(T\), and define
\[
h_T(v)-h_T(u)=
\begin{cases}
1, & e\notin T,\\
-3, & e\in T.
\end{cases}
\]
Around every unit square, exactly one edge is crossed by the domino covering that square, so the total change of height around that square is
\[
1+1+1-3=0.
\]
Since \(P\) is simply connected, this implies that \(h_T\) is well-defined.

Conversely, any function \(h\) on the lattice vertices satisfying the boundary values and such that along every oriented edge the height difference is either \(1\) or \(-3\) determines a domino tiling: put a domino across precisely those edges where the height difference is \(-3\). Around every unit square, the sum of the four height changes is \(0\), and since each height difference is congruent to \(1 \pmod 4\), exactly one edge of each square has height difference \(-3\). Hence each square is paired with exactly one adjacent square, producing a domino tiling.

Thus domino tilings of \(P\) are equivalent to height functions satisfying these conditions.

Now consider a \(2\times 2\) block. A flip replaces two parallel dominoes by the other two parallel dominoes. The four outer vertices of the block have the same heights before and after the flip; only the central vertex changes, by \(4\).

A direct check shows the following:
\begin{itemize}[leftmargin=*]
    \item If the \(2\times 2\) block has top-left square white and is tiled by two vertical dominoes, namely
    \[
    \begin{pmatrix}
    W_k & B_t\\
    B_k & W_t
    \end{pmatrix},
    \]
    then the central height is lower by \(4\) than after flipping the block to two horizontal dominoes.
    \item If the \(2\times 2\) block has top-left square black and is tiled by two horizontal dominoes, namely
    \[
    \begin{pmatrix}
    B_k & W_k\\
    W_t & B_t
    \end{pmatrix},
    \]
    then again the central height is lower by \(4\) than after flipping the block to two vertical dominoes.
\end{itemize}
Therefore the two forbidden configurations are exactly the configurations on which one can perform an upward flip, meaning a flip which increases the height at the central vertex by \(4\).

We now show that there exists exactly one height function admitting no upward flip. Let \(\mathcal H\) be the finite set of all height functions arising from domino tilings of \(P\). Since \(P\) is tileable, \(\mathcal H\neq \varnothing\).

First observe that if \(h_1,h_2\in \mathcal H\), then
\[
h(v)=\max\{h_1(v),h_2(v)\}
\]
also belongs to \(\mathcal H\). Indeed, along an oriented edge \(uv\), both differences
\[
h_1(v)-h_1(u),\qquad h_2(v)-h_2(u)
\]
belong to \(\{1,-3\}\). Hence
\[
-3\le h(v)-h(u)\le 1.
\]
Also all height differences are congruent to \(1\pmod 4\). Therefore
\[
h(v)-h(u)\in \{1,-3\}.
\]
So \(h\) is again a valid height function.

It follows that the pointwise maximum
\[
H(v)=\max_{h\in \mathcal H} h(v)
\]
is itself a height function. This \(H\) is the greatest element of \(\mathcal H\).

We claim that \(H\) has no upward flip. If an upward flip were possible, it would increase the height of one vertex by \(4\) while keeping all edge differences valid, producing another height function \(H'\in \mathcal H\) with \(H'(v)>H(v)\) at that vertex, contradicting the definition of \(H\) as the pointwise maximum.

Thus \(H\) corresponds to a tasteful tiling. This proves existence.

It remains to prove uniqueness. Suppose \(h\in \mathcal H\) has no upward flip. Since \(H\) is greatest, we have \(H(v)\ge h(v)\) for every vertex \(v\). Suppose for contradiction that \(H\ne h\). Then the set
\[
D=\{v: H(v)>h(v)\}
\]
is nonempty. Boundary heights are fixed, so \(D\) contains only interior vertices.

Let \(d(v)=H(v)-h(v)\). All height functions have the same values modulo \(4\), so every positive \(d(v)\) is at least \(4\).

We show that some vertex of \(D\) permits an upward flip for \(h\). Suppose not. Pick \(v\in D\). Since \(v\) is not upward-flippable, there is an incident edge preventing the increase of \(h(v)\) by \(4\). More precisely, either:
\begin{enumerate}[leftmargin=*]
    \item there is an oriented edge \(v\to w\) with \(h(w)-h(v)=-3\), or
    \item there is an oriented edge \(w\to v\) with \(h(v)-h(w)=1\).
\end{enumerate}
In either case, comparing with the corresponding valid edge difference for \(H\) shows that \(d(w)\ge d(v)\). Moreover, in both cases, \(h(w)<h(v)\). Thus from \(v\) we can move to another vertex \(w\in D\) with \(h(w)<h(v)\). Repeating this process gives an infinite sequence of vertices in the finite set \(D\) along which the integer \(h\)-value strictly decreases, impossible.

Hence some vertex of \(D\) must be upward-flippable, contradicting the assumption that \(h\) has no upward flip. Therefore \(h=H\).

So there is exactly one height function with no upward flip, and therefore exactly one tasteful tiling.

Thus, if a chessboard polygon can be tiled by dominoes, then it can be tiled tastefully, and that tasteful tiling is unique.

\paragraph{Scoring interpretation.}
The proof gives a complete existence-and-uniqueness argument for the original theorem using height functions, so full proof credit is appropriate under the rubric. However, the construction-centric record asks for an explicit witness payload, not only a proof that such a tiling exists. Since the submitted construction is not accepted as a complete mechanically checkable witness, full construction-centric credit would overstate the model's \textit{Constructive Realization}. The verifier-gated demotion from \(7\) to \(6\) is therefore appropriate: the proof is complete, but the requested construction object is not successfully realized.

\subsection{Case 3: Proof Score 7 and Construction Score 1}

\begin{table}[h]
\centering
\small
\setlength{\tabcolsep}{3pt}
\begin{tabular}{lp{0.56\columnwidth}}
\toprule
Item & Description \\
\midrule
Problem & \texttt{IMO-2020-P4-33} \\
Run & Gemini-3.1-Pro, run 1 \\
Proof score & \(7/7\) \\
Construction result & \(1/1\) \\
Verifier-gated score & \(7/7\) \\
\bottomrule
\end{tabular}
\caption{Case study with full proof score and accepted construction.}
\label{tab:app-case-study-gemini-proof7-construction1}
\end{table}

\begin{mdframed}[style=mdpurplebox,frametitle={Problem IMO-2020-P4}]
There is an integer \(n>1\). There are \(n^2\) stations on a slope of a mountain, all at different altitudes. Each of two cable car companies, A and B, operates \(k\) cable cars; each cable car provides a transfer from one station to a higher one. The \(k\) cable cars of each company have \(k\) different starting points and \(k\) different finishing points, and a cable car which starts higher also finishes higher. Two stations are linked by a company if one can start from the lower station and reach the higher one by using one or more cars of that company. Determine the smallest positive integer \(k\) for which one can guarantee that two stations are linked by both companies.
\end{mdframed}

\paragraph{Relevant grading guidelines.}
\begin{itemize}[leftmargin=*]
    \item \textit{Partial:} Model each company as a directed graph whose connected components are paths, and derive that a company with \(k\) cars has \(n^2-k\) path components.
    \item \textit{Almost:} Prove the sufficiency of \(k=n^2-n+1\) using the path-component structure and pigeonhole principle, with only a minor counting slip or a lightly underspecified counterexample.
    \item \textit{Full:} Prove both sufficiency and sharpness by giving a valid construction for \(k=n^2-n\) with no common linked pair.
\end{itemize}

\paragraph{Full Gemini-3.1-Pro generation.}\mbox{}

\textbf{Solution to Question 1.}
To find the smallest positive integer \(k\) that guarantees two stations are linked by both companies, we can model the cable car network as a directed graph.

Let the \(n^2\) stations be represented by vertices \(V=\{1,2,\dots,n^2\}\), ordered by their altitudes such that \(1\) is the lowest and \(n^2\) is the highest. For each company, the \(k\) cable cars represent \(k\) directed edges. Since each cable car goes from a lower station to a higher one, there are no cycles in the graph. Furthermore, because each company's \(k\) cable cars have \(k\) distinct starting points and \(k\) distinct finishing points, every vertex in the graph has out-degree at most \(1\) and in-degree at most \(1\).

Consequently, the graph for Company A is a collection of disjoint directed paths. Since there are \(n^2\) vertices and \(k\) edges, the number of connected components for Company A is exactly \(n^2-k\). The same logic applies to Company B, which also has exactly \(n^2-k\) connected components.

Two stations are linked by a company if and only if they belong to the same connected component of that company. To avoid having any pair of stations linked by both companies, no two stations can belong to the same component in Company A and the same component in Company B.

Let \(c=n^2-k\). Company A partitions the stations into \(c\) components, and Company B partitions them into \(c\) components. The maximum number of stations such that no two share both an A-component and a B-component is the total number of pairs of components, which is \(c\times c=c^2\). Because there are \(n^2\) stations in total, we must have
\[
n^2\le c^2 \implies n^2\le (n^2-k)^2.
\]
Since \(c>0\), taking the square root gives
\[
n\le n^2-k \implies k\le n^2-n.
\]

If \(k\ge n^2-n+1\), then \(c=n^2-k\le n-1\), which means the total number of possible component pairs is \(c^2\le (n-1)^2=n^2-2n+1<n^2\). By the Pigeonhole Principle, at least one pair of components must contain two or more stations. These stations will therefore be linked by both Company A and Company B. Thus, \(k=n^2-n+1\) guarantees a common linked pair.

To prove that \(k=n^2-n\) does not guarantee such a pair, we can construct a valid configuration with no common linked pairs. Arrange the \(n^2\) stations into an \(n\times n\) grid, mapping the station at row \(r\) and column \(c\), where \(0\le r\le n-1\) and \(1\le c\le n\), to the altitude \(rn+c\).
\begin{itemize}[leftmargin=*]
    \item Company A operates cars between station \(i\) and \(i+1\) for all \(i\in\{1,2,\dots,n^2\}\) except when \(i\) is a multiple of \(n\). This forms \(n\) paths of length \(n\), the rows of the grid. There are \(n(n-1)=n^2-n\) cars. If \(i<j\), then \(i+1<j+1\), ensuring that a car starting higher also finishes higher.
    \item Company B operates cars between station \(i\) and \(i+n\) for all \(i\in\{1,2,\dots,n^2-n\}\). This forms \(n\) paths of length \(n\), the columns of the grid. There are \(n^2-n\) cars. If \(i<j\), then \(i+n<j+n\), again ensuring that a car starting higher finishes higher.
\end{itemize}

The components of A are the rows, and the components of B are the columns. The intersection of any row and any column contains exactly one station. Thus, no two stations are in the same component for both A and B, meaning no two stations are linked by both companies.

Therefore, the smallest positive integer \(k\) is \(n^2-n+1\).
\[
\boxed{n^2-n+1}
\]

\paragraph{Scoring interpretation.}
This response gives the standard component-count proof for the upper bound and an explicit row-column construction for sharpness. The submitted construction payload is also accepted by the verifier. Therefore both \textit{Rigorous Proof Reasoning} and \textit{Constructive Realization} are present, and the verifier-gated score remains \(7/7\). This positive case shows that the construction check is not an automatic penalty; it confirms full credit when the proof and witness agree.

\subsection{Case 4: Proof Score 1 and Construction Score 0}

\begin{table}[h]
\centering
\small
\setlength{\tabcolsep}{3pt}
\begin{tabular}{lp{0.56\columnwidth}}
\toprule
Item & Description \\
\midrule
Problem & \texttt{USAMO-2020-P4-10} \\
Run & DeepSeek-V4-Pro, run 1 \\
Proof score & \(1/7\) \\
Construction result & \(0/1\) \\
Verifier-gated score & \(1/7\) \\
\bottomrule
\end{tabular}
\caption{Case study with partial proof progress and failed construction.}
\label{tab:app-case-study-deepseek-proof1-construction0}
\end{table}

\begin{mdframed}[style=mdpurplebox,frametitle={Problem USAMO-2020-P4}]
Suppose that \((a_1,b_1),(a_2,b_2),\ldots,(a_{100},b_{100})\) are distinct ordered pairs of nonnegative integers. Let \(N\) denote the number of pairs \((i,j)\) satisfying \(1\le i<j\le 100\) and \(|a_i b_j-a_j b_i|=1\). Determine the largest possible value of \(N\) over all possible choices of the 100 ordered pairs.
\end{mdframed}

\paragraph{Relevant grading guidelines.}
\begin{itemize}[leftmargin=*]
    \item \textit{Partial:} Reformulate the determinant condition geometrically or as a unimodular-pair condition; give a construction producing \(N=197\); or prove a nontrivial upper-bound step such as primitivity of participating vectors.
    \item \textit{Almost:} Prove the induction upper bound \(N\le 2n-3\), with at most a localized gap in the ``at most two'' argument, and give or nearly give the matching construction.
    \item \textit{Full:} Prove the sharp value \(197\) with both the matching construction and complete upper bound.
\end{itemize}

\paragraph{Full DeepSeek-V4-Pro generation.}\mbox{}

\textbf{Solution to Question 1.}
Let the 100 distinct ordered pairs be \((a_1,b_1),\dots,(a_{100},b_{100})\) with \(a_i,b_i\in\mathbb{Z}_{\ge 0}\). Define
\[
N=\bigl|\{(i,j)\mid 1\le i<j\le 100,\;|a_ib_j-a_jb_i|=1\}\bigr|.
\]

\begin{enumerate}[leftmargin=*]
    \item \textbf{Only primitive pairs can contribute.}
    Suppose a pair \((a,b)\) has \(d=\gcd(a,b)>1\). Then for any other pair \((x,y)\), the determinant \(ay-bx\) is a multiple of \(\gcd(a,b)=d\). Hence \(|ay-bx|\ge d>1\). Thus a non-primitive pair cannot form an edge with any other pair. To maximize \(N\), we may restrict our attention to primitive pairs, those with \(\gcd(a,b)=1\). The only primitive pairs on the axes are \((1,0)\) and \((0,1)\).

    \item \textbf{Graph interpretation.}
    Consider the set \(V\) of all primitive pairs \((a,b)\) with \(a,b\ge 0\) and \((a,b)\neq(0,0)\). Connect two vertices by an edge if their determinant is \(\pm1\). This graph is exactly the Stern-Brocot tree, or Farey graph. It is well known that the Stern-Brocot tree is a tree: it is connected and contains no cycles.

    \item \textbf{Upper bound.}
    Choosing any 100 vertices from a tree induces a subgraph which is a forest. A forest on \(n\) vertices has at most \(n-1\) edges. Therefore \(N\le 100-1=99\).

    \item \textbf{Achieving the bound.}
    Take the 100 pairs \((1,0),(1,1),(1,2),\dots,(1,98),(1,99)\). All are distinct, nonnegative, and primitive. For any \(0\le k\le 98\),
    \[
    |1\cdot(k+1)-1\cdot k|=1,
    \]
    so each consecutive pair contributes an edge. There are exactly 99 such edges, and no other pair gives determinant \(\pm1\). Hence \(N=99\) is attainable.
\end{enumerate}
\[
\boxed{99}
\]

\paragraph{Scoring interpretation.}
The response contains a valid local observation: only primitive vectors can participate in determinant-one pairs. This matches a partial-credit item in the rubric, so a proof score of \(1/7\) is appropriate. However, the global argument is incorrect: it treats the relevant graph as a tree and concludes \(N=99\), while the benchmark target requires the sharp value \(197\). The construction payload also fails verification. Under the verifier-gated rule, a failed construction does not further demote scores already in the \(0/1\) range; the final score remains \(1/7\), reflecting limited but nonzero mathematical progress.

\subsection{Case 5: Proof Score 1 and Construction Score 1}

\begin{table}[h]
\centering
\small
\setlength{\tabcolsep}{3pt}
\begin{tabular}{lp{0.56\columnwidth}}
\toprule
Item & Description \\
\midrule
Problem & \texttt{USAMO-2021-P3-81} \\
Run & Kimi-K2.6, run 3 \\
Proof score & \(1/7\) \\
Construction result & \(1/1\) \\
Verifier-gated score & \(1/7\) \\
\bottomrule
\end{tabular}
\caption{Case study with partial proof progress and accepted construction.}
\label{tab:app-case-study-kimi-proof1-construction1}
\end{table}

\begin{mdframed}[style=mdpurplebox,frametitle={Problem USAMO-2021-P3}]
Let \(n\ge 2\) be an integer. An \(n\times n\) board is initially empty. Each minute, one may place stones on an allowed L-shaped tromino region without stones, remove all stones from a full column, or remove all stones from a full row. Determine for which \(n\) it is possible that, after some nonzero number of moves, the board has no stones.
\end{mdframed}

\paragraph{Relevant grading guidelines.}
\begin{itemize}[leftmargin=*]
    \item \textit{Partial:} Establish the correct invariant or polynomial encoding, prove the roots-of-unity obstruction for \(3\nmid n\), or give a correct explicit move sequence returning to empty for \(n=3\) or a clear \(3\times3\) gadget reduction.
    \item \textit{Almost:} Prove the impossibility direction for \(3\nmid n\) with only a minor algebraic gap, and give the construction for multiples of \(3\).
    \item \textit{Full:} Completely prove that the possible values are exactly the multiples of \(3\).
\end{itemize}

\paragraph{Full Kimi-K2.6 generation.}\mbox{}

\textbf{Solution to Question 1.}
We claim that the board can be emptied after a non-zero number of moves if and only if \(3\mid n\).

\textbf{Construction when \(3\mid n\).}
Write \(n=3m\). Partition the rows into blocks
\[
R_k=\{3k,3k+1,3k+2\}\qquad(k=0,1,\dots,m-1)
\]
and partition the columns into blocks
\[
C_\ell=\{3\ell,3\ell+1,3\ell+2\}\qquad(\ell=0,1,\dots,m-1).
\]

Perform the following moves.
\begin{enumerate}[leftmargin=*]
    \item \textbf{First batch of placements.} For every \(k,\ell\in\{0,\dots,m-1\}\), place stones at
    \[
    (3k+2,3\ell),\qquad (3k+1,3\ell+1).
    \]
    These \((n/3)^2\) pairs of moves are pairwise disjoint: for a fixed \(k\), the two placements in block \((k,\ell)\) occupy columns \(3\ell,3\ell+1,3\ell+2\), so different \(\ell\) do not interfere; for different \(k\), the rows are disjoint. After all these placements, row \(3k+1\) contains stones in every column \(0,\dots,n-1\), so row \(3k+1\) is completely full.

    \item \textbf{Clear the middle rows.} For each \(k=0,\dots,m-1\), remove row \(3k+1\). After these clears the only remaining stones are
    \begin{itemize}[leftmargin=*]
        \item in row \(3k\): columns \(3\ell+1\) for all \(\ell\);
        \item in row \(3k+2\): columns \(3\ell\) and \(3\ell+1\) for all \(\ell\).
    \end{itemize}

    \item \textbf{Second batch of placements.} For every \(k,\ell\in\{0,\dots,m-1\}\), place stones at
    \[
    (3k+1,3\ell).
    \]
    These moves are also pairwise disjoint and are legal because row \(3k+1\) is now empty and row \(3k\) is empty at column \(3\ell\). After this batch, columns \(3\ell\) and \(3\ell+1\) are completely full for every \(\ell\):
    \begin{itemize}[leftmargin=*]
        \item column \(3\ell\) receives a stone in each row \(3k\), in each row \(3k+1\), and in each row \(3k+2\);
        \item column \(3\ell+1\) receives a stone in each row \(3k\), in each row \(3k+1\), and in each row \(3k+2\).
    \end{itemize}
    Columns \(3\ell+2\) contain no stones at all, since the only stones ever placed in them were in the rows \(3k+1\) cleared in step 2.

    \item \textbf{Clear the full columns.} For each \(\ell=0,\dots,m-1\), remove column \(3\ell\) and then column \(3\ell+1\).
\end{enumerate}

All stones have been removed, and many moves were performed. Hence \(n=3m\) works.

\textbf{Impossibility when \(3\nmid n\).}
Colour the columns by their index modulo \(3\); call the three colour classes \(C_0,C_1,C_2\). For a configuration let \(N_i\) be the number of stones in \(C_i\).

A placement at \((r,c)\) adds stones to two cells in column \(c\) and one cell in column \(c+1\). Hence it contributes either \((2,1,0)\), \((0,2,1)\), or \((1,0,2)\) to \((N_0,N_1,N_2)\), depending on \(c\bmod 3\). In every case each of the three differences
\[
N_0-N_1,\qquad N_1-N_2,\qquad N_2-N_0
\]
increases by exactly \(1\).

A row clear removes one stone from each column, so it leaves the three differences unchanged. A column clear of a column in \(C_i\) subtracts \(n\) from \(N_i\); thus it changes the three differences by multiples of \(n\).

Since the board starts and ends empty, the total change of each difference is \(0\). Consequently the number \(P\) of placements satisfies
\[
P\equiv 0\pmod n.
\]
But every placement adds three stones, while each clear removes \(n\) stones, so
\[
3P=n(R+C)
\]
for some non-negative integers \(R,C\), the numbers of row and column clears. Hence \(n\mid 3P\). If \(3\nmid n\), then \(\gcd(n,3)=1\), so \(n\mid P\) and the congruence \(P\equiv0\pmod n\) gives no contradiction yet.

Now consider the same column-colouring modulo \(3\). Because \(3\nmid n\), the numbers of columns of the three colours in any single row are not all equal modulo \(3\); in particular a row clear changes the differences \(N_0-N_1\), \(N_1-N_2\), and \(N_2-N_0\) by non-zero constants. Adding the three exact change equations for the differences and using \(3P=n(R+C)\) forces \(3\mid n\). Therefore, if \(3\nmid n\), no non-empty sequence of moves can return the board to the empty state.

Thus the required integers are precisely the multiples of \(3\).
\[
\boxed{3\mid n}
\]

\paragraph{Scoring interpretation.}
The response gives a constructive pattern for the divisible-by-3 case, and the submitted construction payload is accepted by the verifier. This demonstrates successful \textit{Constructive Realization}. However, the impossibility proof is not complete: the invariant argument becomes informal near the end and does not fully establish the required obstruction for all \(3\nmid n\). The proof score therefore remains \(1/7\). This case illustrates that a valid witness should not be treated as a bonus that repairs an incomplete proof; the verifier-gated score stays at \(1/7\).

\subsection{Case 6: Proof Score 0 and Construction Score 0}

\begin{table}[h]
\centering
\small
\setlength{\tabcolsep}{3pt}
\begin{tabular}{lp{0.56\columnwidth}}
\toprule
Item & Description \\
\midrule
Problem & \texttt{IMO-2025-P6-23} \\
Run & Qwen3.6-35B, run 1 \\
Proof score & \(0/7\) \\
Construction result & \(0/1\) \\
Verifier-gated score & \(0/7\) \\
\bottomrule
\end{tabular}
\caption{Case study with no proof credit and failed construction.}
\label{tab:app-case-study-qwen-proof0-construction0}
\end{table}

\begin{mdframed}[style=mdpurplebox,frametitle={Problem IMO-2025-P6}]
Consider a \(2025\times2025\) grid of unit squares. Matilda wishes to place rectangular tiles, possibly of different sizes, such that each side of every tile lies on a grid line and every unit square is covered by at most one tile. Determine the minimum number of tiles needed so that each row and each column has exactly one unit square not covered by any tile.
\end{mdframed}

\paragraph{Relevant grading guidelines.}
\begin{itemize}[leftmargin=*]
    \item \textit{Partial:} Give a valid construction using the intended pattern, observe that uncovered squares form a permutation and introduce the LIS/LDS inequality, or set up the letter-counting lower-bound framework.
    \item \textit{Almost:} Prove the lower bound via the LIS/LDS letter-counting method with only a localized counting slip, and essentially complete the construction.
    \item \textit{Full:} Establish the correct minimum with both the sharp construction and a complete lower bound.
\end{itemize}

\paragraph{Full Qwen3.6-35B generation.}\mbox{}

\textbf{Solution to Question 1.}
Let \(n=2025\). We seek the minimum number of tiles required to cover the grid such that exactly one unit square remains uncovered in each row and column. This configuration of uncovered squares forms a permutation matrix.

\textbf{Construction (Upper Bound).}
We demonstrate that \(2n-2\) tiles are sufficient by placing the uncovered squares on the main diagonal at positions \((i,i)\) for \(i=0,1,\dots,n-1\). The remaining cells are partitioned into two disjoint regions:
\begin{enumerate}[leftmargin=*]
    \item \textbf{Upper Triangle \((r<c)\).} For each column \(c\in\{1,2,\dots,n-1\}\), the cells \((0,c),(1,c),\dots,(c-1,c)\) form a vertical rectangle of size \(c\times 1\). There are \(n-1\) such rectangles.
    \item \textbf{Lower Triangle \((r>c)\).} For each row \(r\in\{1,2,\dots,n-1\}\), the cells \((r,0),(r,1),\dots,(r,r-1)\) form a horizontal rectangle of size \(1\times r\). There are \(n-1\) such rectangles.
\end{enumerate}

These \(2n-2\) rectangles are disjoint and cover all cells except the diagonal holes. Thus, \(2n-2\) tiles are sufficient.

\textbf{Minimality (Lower Bound).}
Analysis of small cases \(n=1,2,3,4\) shows that the minimum number of tiles is consistently \(2n-2\). In the diagonal configuration, the upper and lower triangles are disconnected, and each staircase-shaped region requires at least \(n-1\) rectangles to partition. Any deviation from the diagonal to connect these regions introduces additional internal holes that fragment the free space, preventing a reduction in the total number of tiles. Therefore, the minimum number of tiles is \(2n-2\).

For \(n=2025\), the minimum number of tiles is \(2(2025)-2=4048\).
\[
\boxed{4048}
\]

\paragraph{Scoring interpretation.}
The response selects the wrong target value and does not use the required lower-bound mechanism. Its diagonal construction gives far too many tiles and does not address the sharp construction required by the grading guidelines. The construction payload is likewise rejected by the verifier. Since neither \textit{Rigorous Proof Reasoning} nor \textit{Constructive Realization} is demonstrated, the verifier-gated score remains \(0/7\).

\end{document}